\documentclass[10pt,twocolumn,letterpaper]{article}

\usepackage{cvpr}              

\definecolor{cvprblue}{rgb}{0.21,0.49,0.74}
\usepackage[pagebackref,breaklinks,colorlinks,allcolors=cvprblue]{hyperref}

\def\confName{CVPR}
\def\confYear{2025}

\usepackage{bbding}
\usepackage{multirow}

\usepackage{amsmath, algorithm, algpseudocode}
\newcommand*\samethanks[1][\value{footnote}]{\footnotemark[#1]}
\title{Reconstructing In-the-Wild Open-Vocabulary Human-Object Interactions}

\author{
  Boran Wen\textsuperscript{1,2}~~~~~Dingbang Huang\textsuperscript{1}~~~~~Zichen Zhang\textsuperscript{1}~~~~~Jiahong Zhou\textsuperscript{1}~~~~~Jianbin Deng\textsuperscript{1}~~~~~\\ Jingyu Gong\textsuperscript{2}~~~~~Yulong Chen\textsuperscript{1}\thanks{Corresponding authors.}~~~~~Lizhuang Ma\textsuperscript{1}\samethanks~~~~~Yong-Lu Li\textsuperscript{1,2}\footnotemark[1]\samethanks\\
\small{\textsuperscript{1}Shanghai Jiao Tong University, 
\textsuperscript{2}Shanghai Innovation Institute, 
\textsuperscript{3}East China Normal University}\\
\small\texttt{{\{wenboran, 2331970490, zhoujiahong, 13610214637, deng4a42\}@sjtu.edu.cn }} \\ 
\small\texttt{\{llong\_c, lzma, yonglu\_li\}@sjtu.edu.cn, jygong@cs.ecnu.edu.cn}}

\begin{document}
\maketitle
\begin{abstract}
Reconstructing human-object interactions (HOI) from single images is fundamental in computer vision. 
Existing methods are primarily trained and tested on indoor scenes due to the lack of 3D data, particularly constrained by the object variety, making it challenging to generalize to real-world scenes with a wide range of objects. 
The limitations of previous 3D HOI datasets were primarily due to the difficulty in acquiring 3D object assets. However, with the development of 3D reconstruction from single images, recently it has become possible to reconstruct various objects from 2D HOI images.
We therefore propose a pipeline for annotating fine-grained 3D humans, objects, and their interactions from single images. 
We annotated 2.5k+ 3D HOI assets from existing 2D HOI datasets and built the \textbf{first open-vocabulary in-the-wild 3D HOI dataset Open3DHOI}, to serve as a future test set. 
Moreover, we design a novel Gaussian-HOI optimizer, which efficiently reconstructs the spatial interactions between humans and objects while learning the contact regions.
Besides the 3D HOI reconstruction, we also propose several new tasks for 3D HOI understanding to pave the way for future work. \textit{Data and code will be publicly available at \url{https://wenboran2002.github.io/3dhoi/}}.
\end{abstract}

\section{Introduction}
\label{sec:intro}

Human-Object Interaction (HOI) is an important area in action understanding, with numerous datasets and methods proposed. 
In the 2D HOI domain, large-scale image datasets such as HICO-DET~\cite{hicodet} and HAKE~\cite{hake} have been introduced. 
For 3D HOI, datasets like BEHAVE~\cite{behave} and InterCap~\cite{intercap} have also been proposed to study human interactions with objects in 3D. Despite achieving promising results in open-vocabulary and in-the-wild scenarios within 2D HOI, the 3D HOI field faces challenges in generalizing existing methods to real-world images due to dataset limitations and the lack of 3D open-world HOI data.

In 3D HOI, many datasets have been introduced. Datasets such as BEHAVE~\cite{behave}, InterCap~\cite{intercap}, ImHOI~\cite{imhoi}, and PROX-S~\cite{coins} 
provide multi-view RGBD sequences and 3D annotations in \textit{indoor} scenes. 
Though datasets like WildHOI and 3DIR~\cite{wildhoi,lemon} 
are constructed from in-the-wild images, they contain limited object categories and unreal CAD objects.  
To better understand 3D HOIs and apply them to the real world, we need to collect more realistic and diverse data on interactions with objects. Thus, in this work, we propose a novel 3D HOI annotation method for real-world images of any objects and interactions.

In detail, we built our annotation pipeline on two bases: 
1) Existing 2D HOI datasets provide rich 2D annotations, including bounding boxes, and a wide variety of objects and actions. This diversity creates the potential for reconstructing 3D assets from 2D HOI images.   
2) The development of existing image-based 3D object/human reconstruction techniques. 
We selected images with contact interactions from existing 2D HOI datasets, \eg, HAKE~\cite{hake} and SWIG-HOI~\cite{swig-hoi}. Next, we used InstantMesh~\cite{instantmesh} and OSX~\cite{osx} to reconstruct the objects and the human body respectively, and designed an algorithm for automatically reconstructing \textit{rough} 3D interactions. 
Furthermore, we developed two annotation tools: one for filtering the reconstruction quality and the other for annotating 3D spatial positions. We manually annotated over \textbf{2.5k+} images to create an open-vocabulary, in-the-wild HOI dataset, as the test set for future 3D HOI studies and designed tasks and metrics to evaluate their performance.
It consists of \textbf{370} 3D human-object pairs, \textbf{2,561} objects in \textbf{133} categories, and \textbf{3,671} interactions in \textbf{120} categories. 

\begin{figure}
    \centering
    \includegraphics[width=0.8\linewidth]{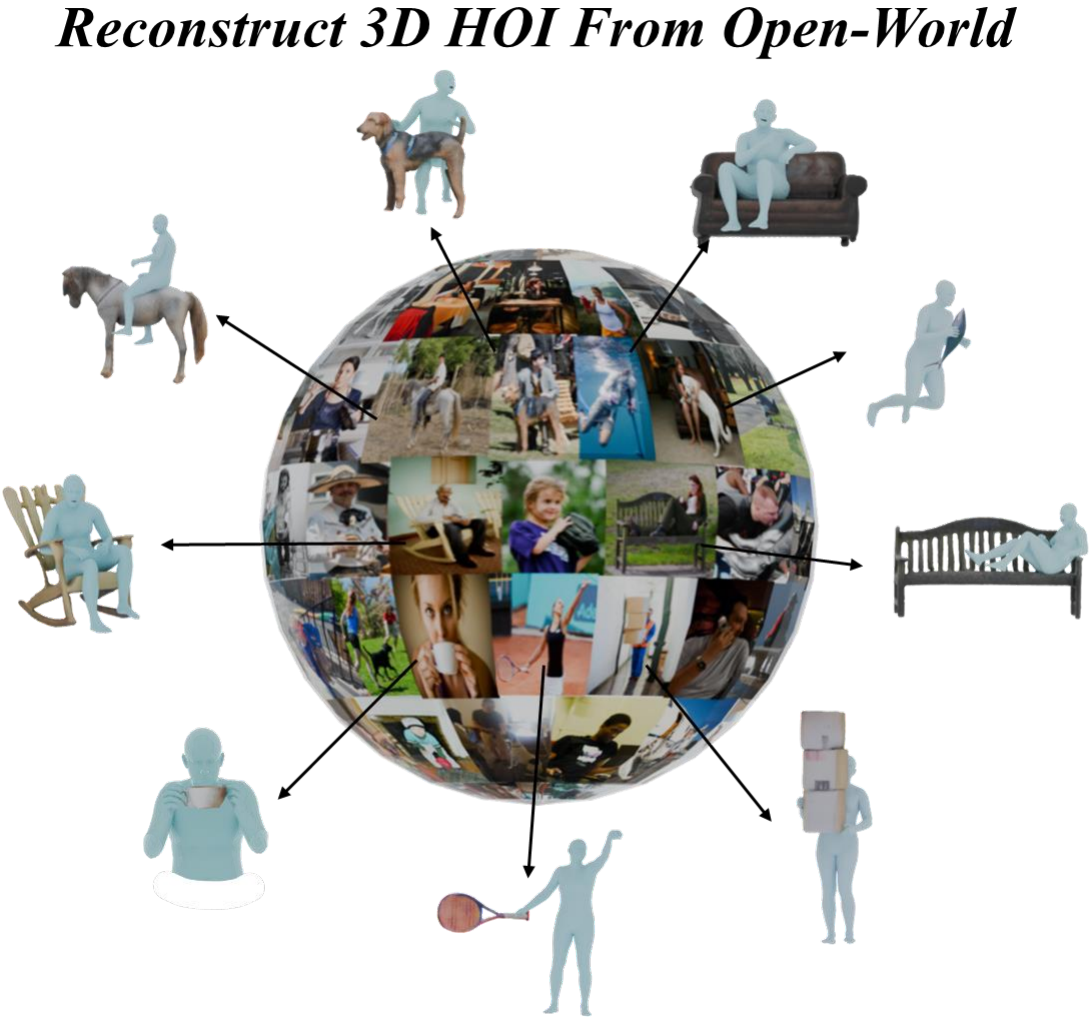} 
    \vspace{-5px}
    \caption{We aim to reconstruct 3D HOIs from arbitrary open-world images. We propose a pipeline for annotating fine-grained reconstructions to build a dataset. Additionally, we introduce a new optimizer suitable for reconstructing arbitrary objects.} 
    \vspace{-10px}
    \label{fig:insight}
\end{figure}

Given the new dataset, we also proposed a \textit{training-free} algorithm for reconstructing 3D HOIs from monocular images.
On one hand, previous training-free algorithms do not require specific object categories or templates, but their optimization performance is generally limited, and they rely on manual annotations. 
On the other hand, previous training-based methods perform well for specific object categories and scenes but struggle to generalize to open-world environments. 
To this end, we leveraged the 3D Gaussian Splatting model to propose a novel Gaussian-HOI optimizer to improve the reconstruction quality. 
It utilizes Gaussian rendering capabilities to ensure that the reconstructed 3D assets are aligned with the image from the target view and takes advantage of the opacity attribute of Gaussians to identify contact regions, which makes the optimization of 3D interaction relationships more effective.

Overall, our contributions are:
1) We utilized SOTA 3D reconstruction tools to develop a 3D HOI annotation method.
2) We built a new and extensive 3D HOI dataset Open3DHOI consisting of 2.5k+ images with rich 2D and 3D annotations.
3) We designed a 3D HOI optimizer based on 3D Gaussian Splatting to reconstruct the spatial interactions between humans and objects from single images.

\section{Related Works}
\subsection{HOI Benchmarks}

The development of 2D HOI benchmarks~\cite{hake,hico-det,v-coco,swig-hoi,hcvrd,gio,djrn,pangea,pastanet} has made our 3D HOI reconstruction approach possible. Datasets like HICO-DET~\cite{hico-det} and HAKE~\cite{hake} provide annotations for 80 object categories and 117 action categories. Additionally, open-vocabulary datasets such as SWIG-HOI~\cite{swig-hoi} include annotations for over 1,000 object categories. 

In contrast, 3D HOI datasets contain significantly fewer action and object categories, and most are recorded in fixed indoor environments. BEHAVE~\cite{behave}, as the earliest one, introduced a method for obtaining accurate 3D HOI annotations from multi-view videos, providing interaction data for 20 common objects. 
InterCap~\cite{intercap} further built upon it by offering more detailed hand interaction information. 
Recently, several benchmarks~\cite{lemon,wildhoi} for reconstructing 3D HOIs from real-world images have been proposed. However, they have notable limitations. Their object categories are focused on a few common types—such as balls, skateboards, and bicycles—and the number of instances is limited, with all objects derived from fixed 3D CAD models.

\subsection{3D Reconstruction}
3D reconstruction has seen rapid advancements recently, both in humans and objects. 
After SMPL~\cite{smpl}, parametric human body modeling has rapidly evolved. Currently, the SMPL-X model~\cite{SMPL-X}, which includes detailed hand and facial expression modeling, is widely used in the field~\cite{osx,smplerx,pymafx2023,cliff}.
We utilized a state-of-the-art one-stage model~\cite{osx} for our human body reconstruction. 
What's more, image-to-3D has emerged as a rapidly advancing area in 3D vision recently. From SDS loss optimization methods~\cite{dreamfusion,magic123} that leverage 2D diffusion priors to Multi-view Diffusion Models~\cite{zero123,one2345,wonder3d, realfusion} and Large Reconstruction Models~\cite{lrm, instant3d, instantmesh} based on large-scale data, the quality and efficiency of 3D generation from a single image have seen significant improvements. 
To generate our 3D HOI dataset, we need to reconstruct a large volume of image data, requiring a \textit{balance} between the quality and efficiency of existing 3D generation methods. Ultimately, we selected InstantMesh~\cite{instantmesh} as our reconstruction model.

\subsection{3D HOI Reconstruction}
Reconstructing 3D HOIs from a single image~\cite{djrn,p3haoi} is a challenging task and important for many applications~\cite{humanvla,imdy,kp}. It requires maintaining consistency between the spatial positions of the human and the object within the image in the given camera view while ensuring that the spatial interactions are realistic and coherent.  
Kanazawa~\etal~\cite{phosa} optimizes spatial interactions through predefined contact pairs, 
while Wang~\etal~\cite{commensense} leverages GPT-3's prior knowledge to optimize spatial interactions. 
Xie~\etal~\cite{chore} learnt HOI spatial arrangement priors from the BEHAVE dataset. Wang~\etal~\cite{wildhoi} learnt the prior distribution of the 2D human-object keypoint layout and viewports to tune the relative pose between the 3D human and object.

\section{3D HOI Annotation}
\label{sec:annotation}
In this section, we introduce our new pipeline for 3D HOI annotation from single-view images.

\subsection{Coarse Reconstruction Annotation}
\label{sec:coarse recon}
First, we used a state-of-the-art human pose estimation method~\cite{osx} to obtain the 3D representation of the human body and employed image-to-3D technique~\cite{instantmesh} to generate the 3D representation of the object. 

\begin{figure}[h]
    \centering
    \includegraphics[width=0.8\linewidth]{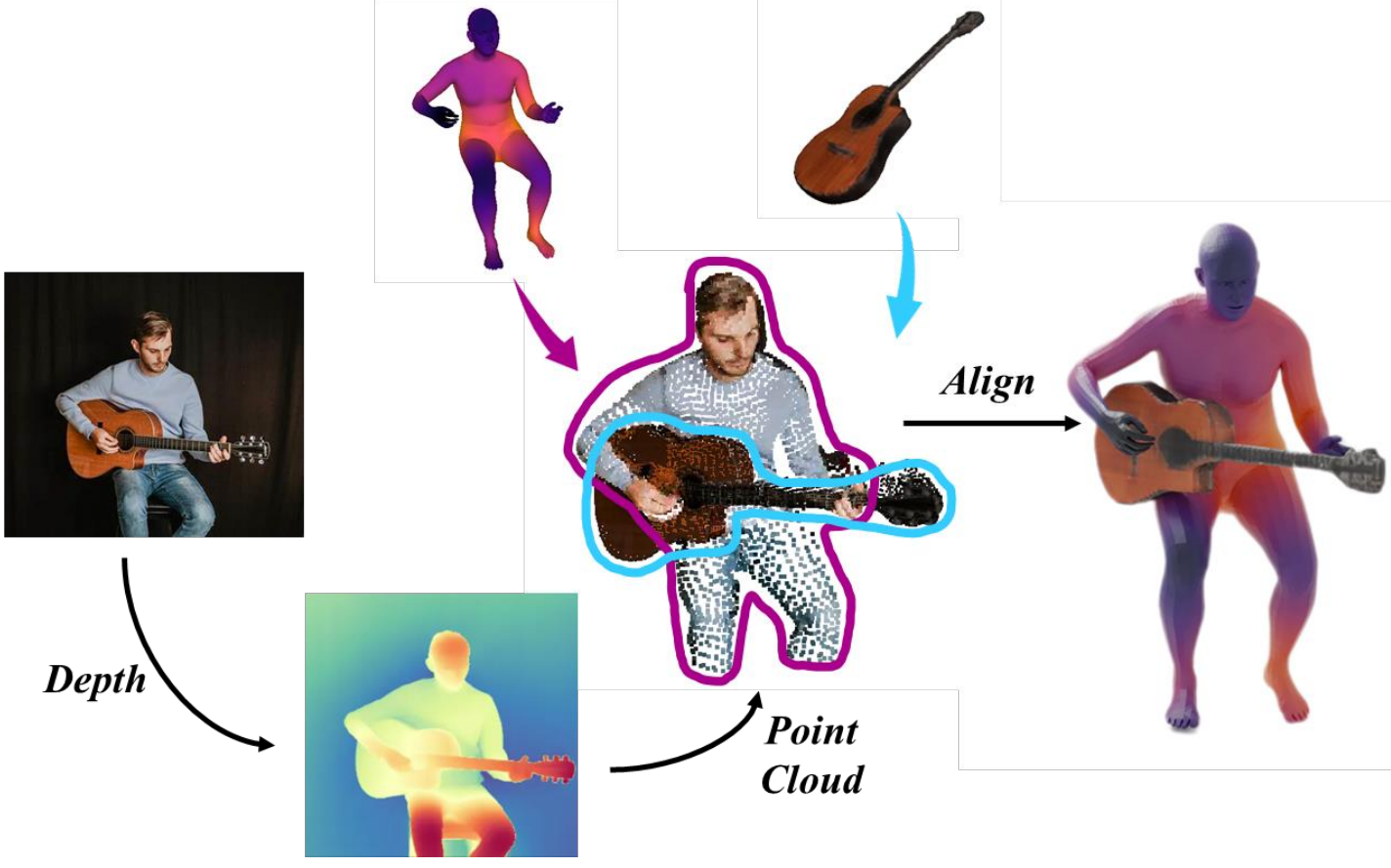} 
    \vspace{-5px}
    \caption{Coarse Reconstruction. We first obtain depth from the images and generate point clouds. Given masks, we extract the corresponding point clouds for the person (pink) and object (blue). We obtain a rough reconstruction by matching the MESH vertices of the person and the object with the depth point cloud.} 
    \vspace{-7px}
    \label{fig:coarse}
\end{figure}

Second, in-the-wild HOI images often involve significant occlusions between humans and objects and between different objects. To address the critical occlusion problem in object 3D reconstruction, we employed occlusion completion~\cite{amodalmask} and used Stable Diffusion 1.5~\cite{sd} as the inpainting tool to obtain complete object images.

Finally, we applied our projection algorithm to estimate the rough spatial relationship, including the positions and sizes of the human and object. We used monocular image depth estimation~\cite{zoedepth} combined with human and object masks, to generate depth point clouds for both the human and the object. 
Then, we obtained a rough estimation of their spatial positions and size by matching the sampled point clouds from the human and object meshes with the depth point clouds as shown in Fig.~\ref{fig:coarse}. 

The rough reconstruction can facilitate the subsequent manual annotation. Rough reconstruction usually has serious mesh collision issues and inaccurate scales and positions. Additionally, the object pose provided by InstantMesh~\cite{instantmesh} is also not accurate. Therefore, to obtain more precise 3D human-object interaction information, we manually performed further annotations.

\subsection{Fine Reconstruction Annotation}
\label{sec:fine annotation}
\begin{figure*}
    \centering
    \includegraphics[width=0.9\linewidth]{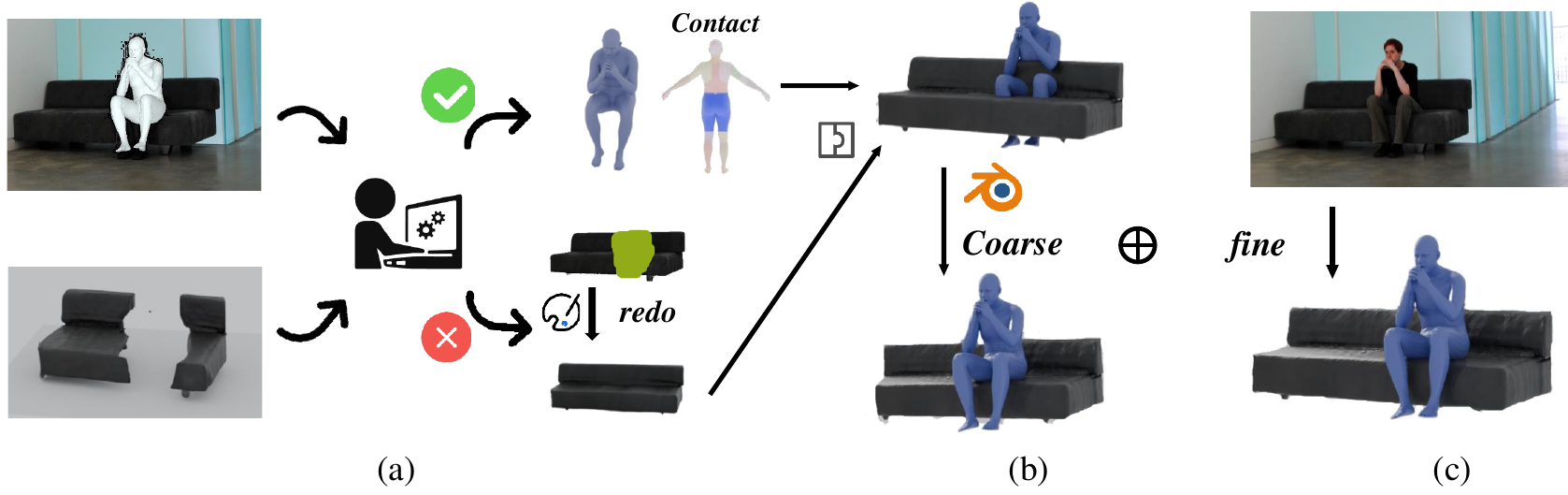}
    \vspace{-5px}
    \caption{Annotation Pipeline. 
    (a) Filtering. Given the reconstructed human and object meshes, annotators assess the quality. If the human reconstruction is eligible, the contact area is further annotated. If the object reconstruction fails, the mask is redrawn manually and the reconstruction is performed again. 
    (b) Given the 3D human interaction through coarse reconstruction, we adjust the object position in Blender. For example, the rough annotation of the couch and the human body shows a mesh collision. We move the object to make sure the person is correctly seated on the couch. 
    (c) We use a fine annotation tool to further align the annotated human and object with the image.} 
    \vspace{-10px}
    \label{fig:manual}
\end{figure*}

We designed two annotation tools to facilitate manual annotating for obtaining more refined 3D HOI information. Fig.~\ref{fig:manual} shows the whole process.

\textbf{Filtering Tool} filters the initial reconstructions and annotates contact regions. 
The filtering consists of:

1) \textbf{Filtering the SMPL-X human reconstruction}. 
We project the reconstructed SMPL-X mesh onto the image to assess whether the pose estimation of key interaction joints is accurate. For joints not involved in the interaction, we relax the criteria. For example, if a person is drinking water and the target object is a cup, we focus on the accuracy of the hand pose reconstruction, while only ensuring that the lower body pose is reasonable.

2) \textbf{Filtering the object reconstruction}. 
Given the reconstructed object mesh, we project it from six viewpoints. Annotators evaluate the quality of the object mesh, retaining images with high-quality reconstructions, especially paying attention to the quality of the interaction area.

3) \textbf{Manual optimization of the inpainting mask}. 
The inpainting masks obtained from occlusion completion fail sometimes. Annotators can manually correct them using a brush, ensuring higher accuracy. In Fig.~\ref{fig:manual}, (a) shows that the annotator manually drew a mask (green part) on the poorly reconstructed couch. After re-inpainting and reconstruction, it achieves a much better result. 
Considering that some objects may only partially appear in the image, we still allow for the completion of the object by manually editing the mask. Annotators can use a brush to fill in the parts of the object that are outside the image. 
We further divided the SMPL-X mesh into 34 human body regions, which are used for contact area annotations. Annotating contact areas at a fine-grained, vertex level would be costly and difficult to ensure high-quality results. Therefore, we opted for \textbf{part-level} contact annotations, which are sufficient for most interaction scenarios. In Fig.~\ref{fig:manual}, (a) shows the annotation of the contact area, where the person is sitting on the couch, and the thighs and bottom are annotated (blue part).

\begin{figure}
    \centering
    \includegraphics[width=0.9\linewidth]{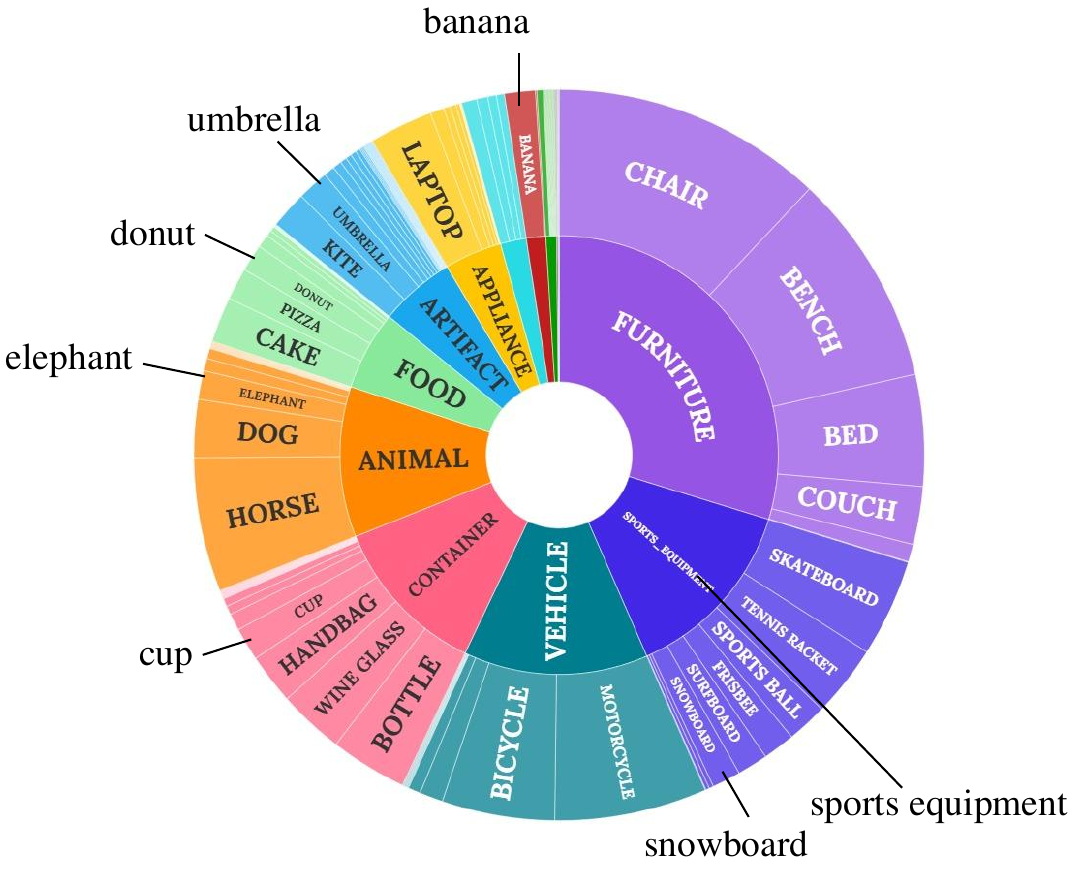} 
    \vspace{-5px}
    \caption{Object category distribution in Open3DHOI. It encompasses a wide range of object categories.} 
    \vspace{-10px}
    \label{fig:dataset}
\end{figure}

\textbf{3D Interaction Tool}
is designed for annotating the spatial interaction between humans and objects. We adopted a \textit{coarse-to-fine} approach. 
Initially, we developed a 3D HOI annotation tool via Blender, where volunteers can adjust the objects' positions, rotations, and scales using a mouse. In Blender, the annotations cannot be compared pixel-by-pixel with the images, and the focus is more on the 3D interaction quality. 
Thus, we further performed detailed annotations for cases where there is a significant discrepancy when projected on the images. We built a website annotation tool based on ImageNet3D~\cite{imagenet3d}. In this refine-annotation tool, volunteers can finetune the annotations by clicking buttons to translate, rotate, and scale the objects based on the projection of the object on the image. 
To ensure the accuracy of the 3D interactions, we provide multiple viewpoints projections of humans and objects, allowing them to more accurately evaluate and adjust the annotations.

\subsection{Open3DHOI Dataset Construction}
We selected 12k+ images from HAKE-Large and 3k+ images from SWIG-HOI, totaling 15k+ images as the database for our 3D reconstruction.  We also collected some images from the website. 
After manual annotation and filtering, we obtained \textbf{2.5k+} images to form our dataset \textbf{Open3DHOI}. 
To our best knowledge, this is the \textbf{first in-the-wild, open-vocabulary 3D HOI dataset based on real-world images}.

\begin{table}[h]
\centering
\resizebox{\linewidth}{!}{
\begin{tabular}{l cc cc cc cc}
\hline
Datasets & Objects &  Action & Human & Object Pose & Contact & 2D HOI\\ 
\hline
BEHAVE~\cite{behave} & 10  & N/A & SMPL+H & \Checkmark & \Checkmark & \XSolidBrush\\ 
InterCAP~\cite{intercap} & 10  & N/A & SMPL-X & \Checkmark & \Checkmark & \XSolidBrush\\
WildHOI~\cite{wildhoi} & 8 & N/A & SMPL &\Checkmark & \XSolidBrush & \XSolidBrush\\
3DIR~\cite{lemon} & 21  & 17 & SMPL+H & \XSolidBrush & \Checkmark & \Checkmark\\
PROX-S~\cite{coins} & 40  & 17 & SMPL-X & \Checkmark & \Checkmark & \XSolidBrush\\
 \hline
Ours& \textbf{133} & \textbf{120} & SMPL-X & \Checkmark & \Checkmark & \Checkmark\\
 \hline
\end{tabular}}
\vspace{-5px}
\caption{Dataset comparison between previous datasets and ours.}
\label{tab:dataset_compare}
\vspace{-7px}
\end{table}

The 2D annotations include bounding boxes, HOI triplets, object labels, and masks obtained using SAM~\cite{sam}. 
The 3D annotations consist of object meshes, SMPL-X parameters for humans, the 6D poses of both objects and humans in space, as well as human contact regions. 
Our dataset includes \textbf{133} object categories and more than \textbf{120} interactions, significantly surpassing the current benchmarks in terms of semantic diversity. We referred to WordNet's classification of object categories and divided our object categories into several major categories, as shown in Fig.~\ref{fig:dataset}. It can be seen that our data includes a wide range of object categories, many of which were rarely attempted in previous 3D HOI datasets, such as food and animals.

We compared the differences between existing 3D HOI datasets and our dataset in Tab.~\ref{tab:dataset_compare}. It can be seen that the object and action categories in the current benchmarks are much fewer than those in our dataset. At the same time, we also provide more detailed 2D and 3D annotations.

\section{Method} 

\subsection{3D Gaussian Splatting}
3D Gaussian Splatting~\cite{kerbl3Dgaussians} is advanced for rendering and reconstructing scenes by representing objects continuously, in a volumetric manner. 
Instead of relying on traditional mesh-based models, it leverages a collection of Gaussian kernels, each defined by its mean, covariance, and intensity, to describe the spatial distribution of objects in 3D. These Gaussian splats, which are soft, overlapping volumetric primitives, enable high-quality rendering while maintaining flexibility in representation. 
For a given pixel $x$, the depth of each overlapping 3D Gaussian is computed using the viewing transformation $W$, resulting in a depth-sorted list of Gaussians $N$. The final color of the pixel is then determined using alpha compositing, expressed as:
\begin{eqnarray}
C=\sum_{n=1}^{|\mathcal{N}|} c_n \alpha_n^{\prime} \prod_{j=1}^{n-1}\left(1-\alpha_j^{\prime}\right),
\label{eq:eq1}
\end{eqnarray}
where $c_n$ denotes the color associated with the 
$n-th$ Gaussian. The effective opacity $\alpha_j^{\prime}$ is got by multiplying the learned opacity $\alpha_n$ by a Gaussian weighting function:
\begin{eqnarray}
\alpha_n^{\prime}=\alpha_n  \exp \left(-\frac{1}{2}\left(\boldsymbol{x}^{\prime}-\boldsymbol{\mu}_n^{\prime}\right)^{\top} \boldsymbol{\Sigma}_n^{\prime-1}\left(\boldsymbol{x}^{\prime}-\boldsymbol{\mu}_n^{\prime}\right)\right).
\end{eqnarray}
Here, $x^\prime$ is the pixel's projected coordinate and ${\mu}_n^{\prime}$ is the projected center of the $n-th$ Gaussian.
Recent works~\cite{hu2023gauhuman,lin2022occlusionfusion} have applied 3D Gaussian Splatting to high-quality human reconstruction, which not only ensures rendering quality but also significantly improves reconstruction speed. For example, in GauGAN-based approaches like Gauhuman~\cite{hu2023gauhuman}, SMPL's vertices are used as the initial point clouds for 3D Gaussian splatting. This method simultaneously learns the SMPL pose and Linear Blend Skinning (LBS) parameters to optimize the human Gaussians, ultimately achieving high-quality reconstruction results. It successfully demonstrates the effectiveness of 3D Gaussian splatting for optimizing human reconstruction based on SMPL representations. Also, work like GS-pose~\cite{gspose} and 6D-GS~\cite{6dgs} leverage 3D Gaussian splatting to optimize the 6D pose of objects. We reasonably infer that this approach can also be effectively applied to reconstruct human-object interactions from a single viewpoint, potentially replacing traditional silhouette-based optimizers~\cite{phosa}.

\subsection{HOI-Gaussian Optimizer}

\begin{figure*}
    \centering
    \includegraphics[width=0.95\linewidth]{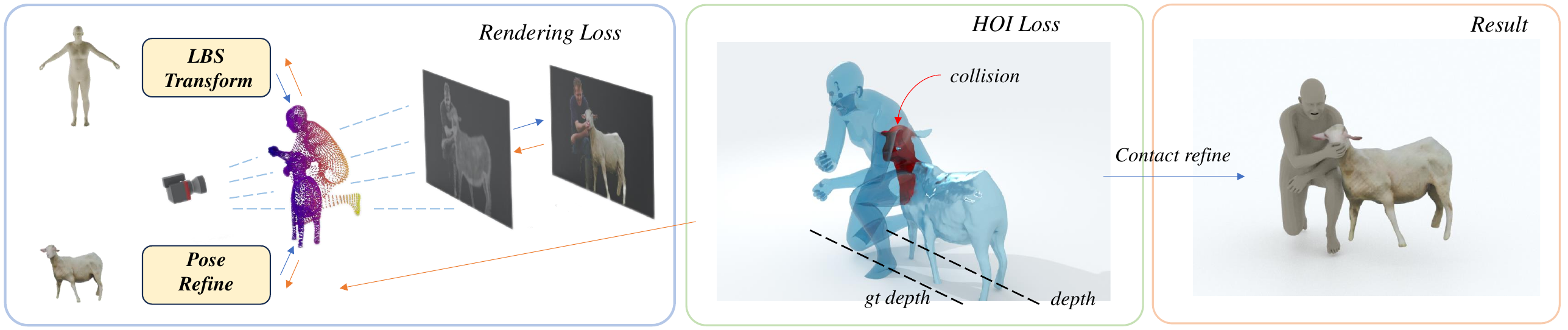}
    \vspace{-5px}
    \caption{Our pipeline. The optimizer first converts the human and object into 3D Gaussian points, then calculates a rendering loss by comparing the Gaussian-rendered image with the ground truth image. This loss is backpropagated to update the object’s pose parameters and the human’s LBS parameters. We also calculate an HOI loss, which includes collision, depth and contact losses, the red overlapping areas between the human and object in the image represent collision regions and the dashed lines represent the ground truth depth and the depth during the optimization process. Finally, we refine the result by optimizing the contact regions.} 
    \vspace{-15px}
    \label{fig:pipeline}
\end{figure*}

We developed the HOI-Gaussian optimizer specifically for 3D HOI reconstruction based on Gauhuman as shown in Fig.~\ref{fig:pipeline}. We chose 3D Gaussian over other silhouette-based optimization methods as we believe it offers the following advantages: 

\textbf{1)} Methods like Gauhuman have demonstrated that 3D Gaussian can be used to adjust human body parameters. Our HOI-Gaussian optimizer can simultaneously optimize object and human poses beyond traditional methods. 

\textbf{2)} 3D Gaussian uses depth from point clouds better to align with the image, reducing cases where large pose discrepancies occur despite small silhouette losses. 

\textbf{3)} We hope to use the features rendered from the 3D Gaussian point clouds to obtain potential contacts, thus reducing reliance on prior, such as manually annotated human-object parts pairs introduced by methods like PHOSA.

We used the vertices from the SMPL-X model to initialize the human 3D Gaussians $g_h$, and the vertices from the object mesh to initialize the object 3D Gaussians $g_o$. During optimizing, we follow GauHuman's pose refinement and LBS offset to learn the parameters of the human model, while introducing a learnable parameter \( W_{obj} \) to optimize the object's 6D pose. Our final interaction 3D Gaussian \( g_{hoi} \) is derived from \( g_h \) and \( g_o \) through
\vspace{-5px}
\begin{eqnarray}
\begin{aligned}
& p^{\text {hoi }}=p^h \oplus p^o, 
\quad \Sigma^{hoi}=\Sigma^h \oplus \Sigma^o, \\
& p^h=R^h p^{h^{\prime}}+t^h, 
\quad p^o=R^o s^o p^{o^{\prime}}+t^o, \\
& \Sigma^{h}=R^h \Sigma^{h^{\prime}} R^{h^{\top}},
\quad \Sigma^o=R^o \Sigma^{o^{\prime}} R^{o \top},
\end{aligned}
\vspace{-10px}
\end{eqnarray}
where \( p \) represents the 3D position and \( \Sigma \) represents the covariance matrix. \( R^h \) and \( t^h \) are the rotation matrix and translation vector obtained from the SMPL-X model through pose and LBS parameters. \( R^o \), \( s^o \), and \( t^o \) represent the object's rotation matrix, scale factor, and translation vector, respectively, and all three variables are learnable. 

\subsection{Contact in Gaussian Model}
The rendering capability of Gaussians ensures that the reconstructed human-object interactions are consistent with the images, while the depth information and rendering characteristics of Gaussians enable us to obtain potential contact areas. 
Given that in monocular images it is difficult to directly determine the contact area between the human and the object, however, we can identify areas where there is no interaction easily, \ie, regions in the image where the human and object \textit{do not occlude each other}. 
This allows us to infer potential contact areas between the human and object. 

\begin{figure}
    \centering
    \includegraphics[width=\linewidth]{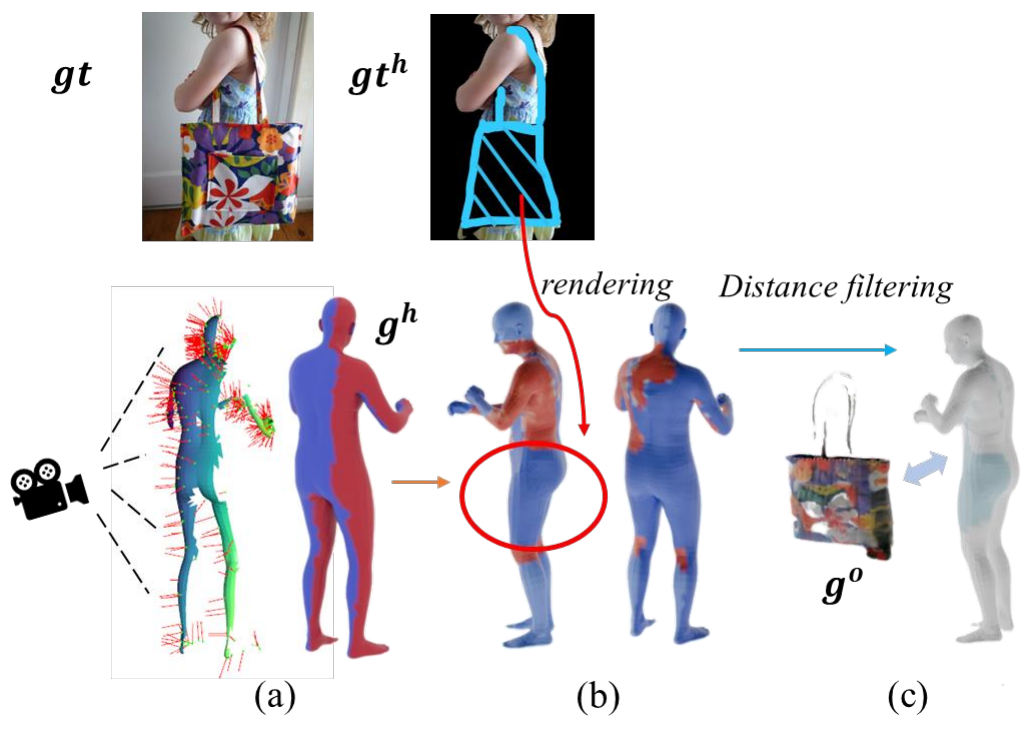}
    \vspace{-10px}
    \caption{Contact region. 
    (a) Opacity initialization using human normals.
    (b) The distribution of human body point cloud opacity scores is visualized to identify the \textit{blue} region as a potential interaction area.
    (c) Based on the approximate distance between the human body and the object, the optimized contact region is further identified, shown in \textit{light blue}. }
    \vspace{-10px}
    \label{fig:contact_expl}
\end{figure}

In the optimization of \( g^{hoi} \), the Gaussian points where \( g^h \) and \( g^o \) occlude each other tend to have \textit{lower opacity} $\alpha$. 
In Eq.~\ref{eq:eq1}, the color of a pixel is influenced by the opacity of the Gaussian points. If the opacity is very low, the contribution of that Gaussian point to the pixel's color becomes less. While optimizing \( g^{hoi} \), we also simultaneously optimize \( g^h \) and \( g^o \). 
Given the original 2D image $gt$ and human image \( {gt}^h \) (Fig.~\ref{fig:contact_expl}), the  occluded areas between the human and the object appear as background in \( {gt}^h \), so does the object image \( {gt}^o \).
Thus, the opacity of the Gaussian points projected in these areas will tend to decrease in rendering.

Additionally, even in regions without overlap in 2D, occlusion relations exist in 3D depth.
Relative to the camera, we consider points in the back to be potential contact areas. 
Without constraints, both front and back points relative to the camera would participate in rendering. 
Thus, we set a very low opacity for points facing back from the camera by calculating normals in advance when initializing Gaussians. This approach is based on the assumption that the human pose will not undergo significant changes in optimization, effectively resolving the issue. Setting a low initial opacity allows points closer to the camera to be prioritized in rendering. As the Gaussian point scale increases, it naturally occludes points further back, preventing them from contributing to the rendering. 
In Fig.~\ref{fig:contact_expl} (a) and (b), the \textit{blue} area represents regions with high opacity, while the \textit{red} area starts with relatively high opacity. As the optimization progresses, occluded parts of the human leg turn blue, designating them as potential contact regions. Meanwhile, the opacity of points behind the camera remains relatively stable.

Additionally, in the optimization, we will have a relatively stable positional relationship between the human and the object before we use the contact region to further optimize. By setting a distance threshold between human and object, we can further narrow down the potential contact area, shown in Fig.~\ref{fig:contact_expl} (c). Therefore, we introduce a new attribute \( c \) into \( g^{h} \), which represents the contact interaction score of a Gaussian point. The calculation of \( c \) is given by 
\vspace{-5px}
\begin{equation}
c=w_{\alpha} \cdot {Norm}(\alpha^h)+w_d \cdot d_C(p^h, p^o)^h,
\end{equation}
where ${Norm}$ means the normalization of the vector to a range between 0 and 1, and \( d_C \) means the Chamfer distance.

\subsection{Loss Function}
Our loss function is divided into two parts: Gaussian rendering loss which is used to optimize the 2D alignment, and HOI loss which is used to optimize the spatial interaction between the human and the object. 

\textbf{Rendering Loss.} 
We adopt the training loss used in 3D Gaussian Splatting, including the L1 loss between the rendered image and GT image, the L2 loss between the rendered mask and GT mask, as well as SSIM (Structural SIMilarity index) loss and LPIPS (Learned Perceptual Image Patch Similarity) loss. 
To ensure the rendering quality of both the human and the object individually, as well as their combined rendering, we perform separate rendering and loss calculations for \( g^{hoi} \), \( g^h \), and \( g^o \):
\vspace{-5px}
\begin{equation}
\begin{aligned}
& \mathcal{L}_r=w_h \mathcal{L}_r^h+w_o  \mathcal{L}_r^o+w_{hoi} \mathcal{L}_r^{\text {hoi }}. \\
\end{aligned}
\vspace{-3px}
\end{equation}

\textbf{HOI Loss}.
We use the HOI loss to constrain the spatial interaction between the human and the object, ensuring its plausibility. 
First, we calculate the Chamfer distance between the human contact area, obtained through the Gaussian rendering process, and the object as a contact loss. At the same time, to ensure that the human and object meshes do not intersect, we follow \cite{collision} to add a collision loss $\mathcal{L}_{colli}$. We also follow \cite{phosa} by adopting an Ordinal Depth loss $\mathcal{L}_{depth}$ to constrain the depth relationships. Our final HOI loss and the total training loss are: 
\vspace{-5px}
\begin{equation}
\begin{aligned}
\mathcal{L}_{{hoi}} = \mathcal{L}_{{cont}}  + \mathcal{L}_{{colli}} + \mathcal{L}_{{depth}},\\
\mathcal{L} = w_r \cdot \mathcal{L}_{{r}} + w_{hoi} \cdot \mathcal{L}_{{hoi}}.
\end{aligned}
\vspace{-3px}
\end{equation}

\section{Experiments}

\begin{figure}[t]
    \centering
    \includegraphics[width=0.8\linewidth]{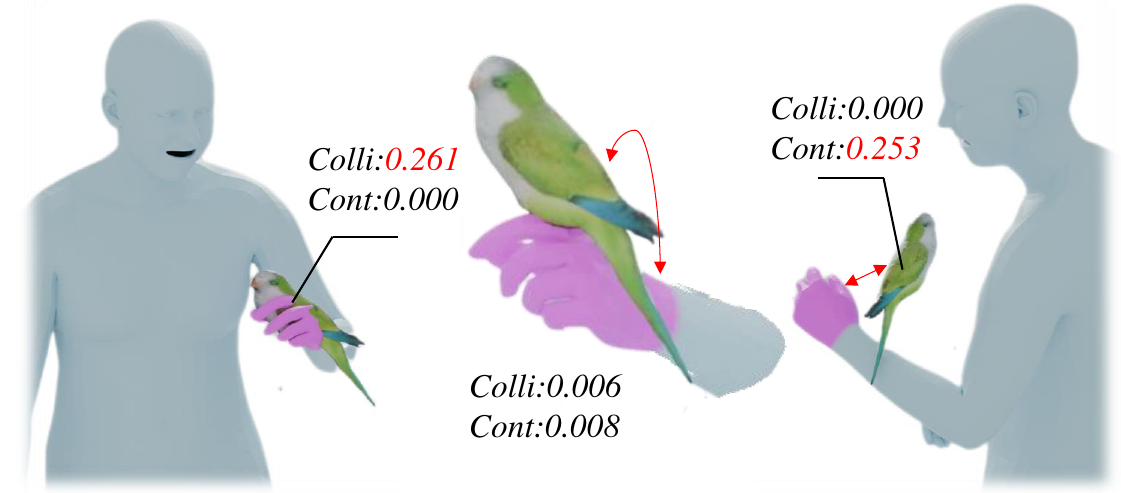}
    \vspace{-3px}
    \caption{Visualization of our ${Co}^2$ Metric. For example, the Colli score focuses on the collision between the bird and the person, while the Cont score calculates the mean Chamfer Distance between the annotated contact part (left hand) and the bird. } 
    \vspace{-7px}
    \label{fig:metrics}
\end{figure}

\subsection{Implementation Details}
Experiments are conducted on our Open3DHOI test set and our method doesn't need training as an optimizer. Object meshes with pose before manual annotation are given. We adopt a staged optimization, optimizing each image for 160 iterations. For the first 100 iterations, only the rendering loss is optimized, and HOI loss is added in the subsequent iterations. Due to our dataset being open-vocabulary, it would be unfair to compare methods trained for specific objects on our dataset, so we use the same training-free method PHOSA~\cite{phosa} as our baseline. To ensure fairness, both our method and PHOSA initialize human body parameters using the parameters from our dataset. 

\begin{table}[h]
\centering
\resizebox{0.85\linewidth}{!}{
\begin{tabular}{l c c c c}
\hline 
 Methods & Scale\textdownarrow & Translation (cm)\textdownarrow & Rotation\textdownarrow  & Cf Distance(cm) \\ \hline
 PHOSA\cite{phosa} & 0.39 & 77.79 & 0.95 & 49.1 \\
 Ours w/o HOI Loss & 0.25 & 38.66 & 0.45  & \textbf{16.9}\\ 
 Ours  & \textbf{0.16} &\textbf{38.44} & \textbf{0.41}  & 19.3 \\ 
 \hline
\end{tabular}}
\vspace{-5px}
\caption{Comparison on object pose metrics.}
\label{tab:comparison}
\vspace{-10px}
\end{table}

\begin{table}[h]
\centering
\resizebox{0.8\linewidth}{!}{
\begin{tabular}{c c c c }
\hline 
 Methods & ${Co}^2$\textdownarrow & Collision\textdownarrow & Contact\textdownarrow  \\ \hline
 PHOSA~\cite{phosa} & 0.431 & 0.105 & 0.326  \\
 Coarse Recon & 0.248 & 0.083 & 0.165 \\
 Ours Gs only  & 0.287 & 0.136 & 0.151  \\
Gs\& depth & 0.216 & 0.080 &0.136 \\
Gs\& colli & 0.189 & 0.046 & 0.143 \\
Gs \& depth \& colli & 0.188 & \textbf{0.045} &0.143 \\
Gs \& depth \& colli \&cont & \textbf{0.181} & 0.053 & \textbf{0.128} \\
 \hline
\end{tabular}}
\vspace{-5px}
\caption{Comparison on collision and contact metrics.}
\label{tab:scores}
\vspace{-10px}
\end{table}

\begin{table}[h]
\centering
\resizebox{0.67\linewidth}{!}{
\begin{tabular}{l c c c c c}
\toprule
\multirow{2}{*}{score}&  \multicolumn{2}{c}{Action}  & \multicolumn{2}{c}{Object} \\ \cline{2-5} & obj & w/o obj &action & w/o action \\ \midrule
 Top-1& 0.47 & 0.20 & 0.32& 0.31\\ 

\bottomrule
\end{tabular}}
\vspace{-5px}
 \caption{Top-1 Accuracy under different prompts.}
 \label{tab:pointllm}
 \vspace{-7px}
\end{table}

\subsection{Metrics}
To more accurately evaluate the reconstruction quality, we used two metrics. 
First, we compared the reconstructions with the object pose in our annotated data, evaluating the differences in scale, translation, rotation and Chamfer Distance. The scale measures the difference between the predicted and GT objects' sizes. The translation is the distance between the predicted and GT objects in \textit{cm}. Rotation is the norm of the predicted rotation matrix and the eye matrix. We designed an alternative metric that better evaluates 3D interaction quality. It combines the extent of collision between the human and the object with the Chamfer distance between the human and the object within our annotated contact regions. We call this metric as \textbf{${C_o}^2$ (Collision-Contact) score}:
\vspace{-10px}
\begin{equation}
    C_o^2=\operatorname{Sig}(\operatorname{Colli}(h,o))+\operatorname{Sig}(\sum_i^p(\operatorname{Cont}(i) / Size )),
\vspace{-5px}
\end{equation}
where $Sig$ is sigmoid function, ${Colli}(h,o)$ is the collision between human mesh and object mesh. $Cont$ calculates the Chamfer distance between each human body part and the object. $Size$ is the object mesh size.

\subsection{Analysis}

\textbf{Gaussian Advantages}. The results show that our method significantly outperforms PHOSA. Our method achieves a higher score in Rotation compared to PHOSA because PHOSA optimizes object pose solely through silhouette loss. In contrast, the 3D Gaussian approach can utilize color matching and richer features, reducing cases where there is minimal silhouette difference but significant disparity from the image. Additionally, our method with contact optimization improved the ${Co}^2$ score, particularly the Contact score. This indicates that the contact regions derived from Gaussian depth information effectively enhanced 3D human-object interaction quality.
\begin{figure}
    \centering
    \includegraphics[width=0.9\linewidth]{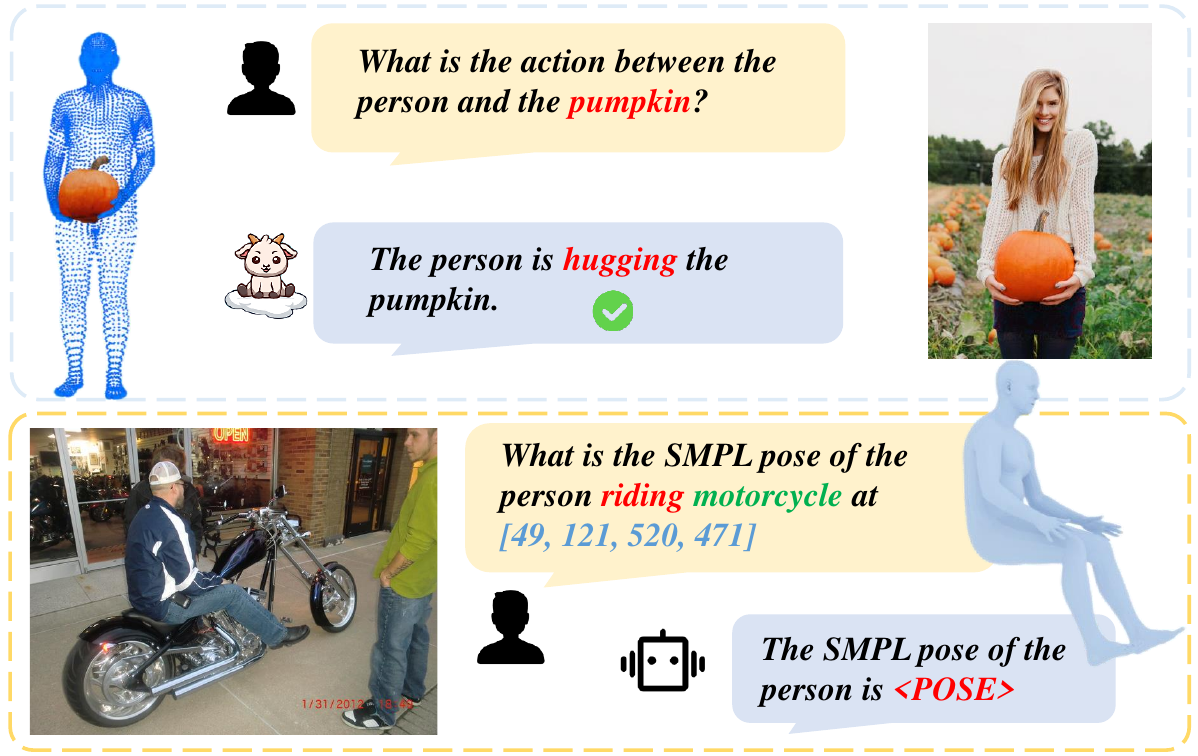}
    \vspace{-5px}
    \caption{Visualization of HOI Understanding and HOI Pose Chat.} 
    \vspace{-15px}
    \label{fig:other_tasks}
\end{figure}

\textbf{Ablation Study}. 
Coarse Recon in Tab.~\ref{tab:scores} is the coarse reconstruction using depth and projection in \ref{sec:coarse recon}. The ${Co}^2$ score using only Gaussian optimization is lower than that of Coarse Recon. Since Gaussian optimization alone does not greatly enhance spatial interaction information, it mainly refines object pose according to the image. However, after adding depth, collision, and contact losses, the 3D score improves significantly, demonstrating that HOI losses are highly effective in optimizing interactions. After adding the contact loss, the collision score slightly decreased, but the contact score improved significantly. Because optimizing the contact area sometimes increases collision in certain images, as objects are moved closer to the intended contact regions. ${Co}^2$ metric is to better balance this trade-off.

\begin{figure*}
    \centering
    \includegraphics[width=0.9\linewidth]{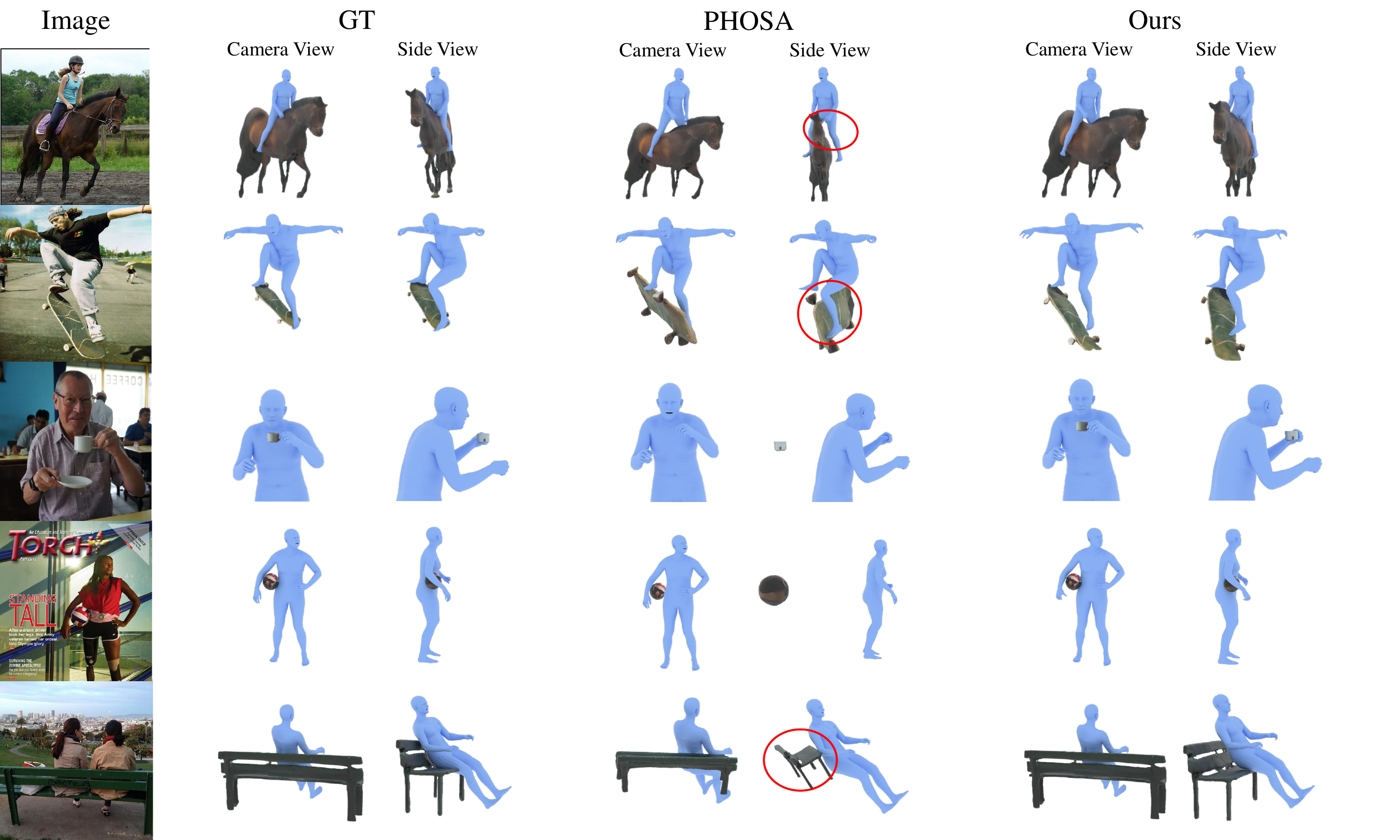}
    \vspace{-5px}
    \caption{Visualized results comparison between GT, PHOSA, and Ours.} 
    \vspace{-12px}
    \label{fig:result}
\end{figure*}
\section{More Tasks}
Our dataset, with its extensive 2D and 3D annotations, can be utilized for various other tasks. In this section, we propose two more tasks as shown in Fig.~\ref{fig:other_tasks}.

\subsection{3D HOI Understanding} 
Understanding 3D assets has been a long-standing area of interest, and recently some large models~\cite{pointllm, 3dllm, GPT4Point} have achieved impressive results in 3D object comprehension. We tested the current state-of-the-art point clouds understanding model PointLLM~\cite{pointllm} on our 3D HOI data to evaluate its capability in action understanding. 
We provide the model with the point clouds of annotated human and object and ask it to answer the interaction verb between the human and the object.  We then compared the output action with GT action annotation. We ask the LLM ``What is the action between the person and the [object]?'' and ``What is the person [Interacting] with?'', where [object] and [Interacting] can be replaced with specific verbs and object or not.

We used the Top-1 score as a metric, the results in Tab.~\ref{tab:pointllm} indicate that PointLLM demonstrates a certain level of understanding of human interaction point clouds, though it remains limited. Given object name significantly improves action answering performance because PointLLM will estimate the action according to the object category with common sense, but given action name will not improve object answering score, which indicates that PointLLM has limited ability to understand the interaction in the point cloud.

\subsection{HOI Pose Chat}
Recently, large models have focused on integrating semantics with 3D data. ChatPose~\cite{chatpose} uses a framework with LLMs to understand and infer 3D human poses from images or textual descriptions. 
F-HOI~\cite{fhoi} leverages large models to unify various HOI tasks. Since our dataset provides 2D annotations of HOI semantics, we tested the open-sourced large model ChatPose. To evaluate its HOI reasoning and pose generation, we selected cases from the dataset with more than one person, provided the model with the target object's location, and asked it to output the SMPL pose of the person interacting with that object. We then compared this generated SMPL pose with the GT SMPL pose from our dataset. We ask Chatpose ``What is the SMPL pose of the person [Interacting] [object] at [Location]?''.

The results in Tab.~\ref{tab:chatpose} indicate that ChatPose's ability to accurately locate the target human body and obtain the correct pose still needs improvement. In the future, we hope our dataset can help drive the development of more powerful models capable of better understanding images and simultaneously obtaining both human and object poses.

\begin{table}[h]
\centering
\resizebox{0.6\linewidth}{!}{
\begin{tabular}{c c c }
\hline 
 Prompt&  MPJPE\textdownarrow &  MPVPE \textdownarrow \\ \hline
 Action & 103.6 & 131.2 \\ 
 w/o Action & 105.2 & 133.5  \\
 Action + Object & \textbf{103.4} & \textbf{130.9} \\ \hline
\end{tabular}}
\vspace{-5px}
 \caption{ChatPose performance results under different prompts.}
 \label{tab:chatpose}
 \vspace{-10px}
\end{table}

\section{Discussion} 

We propose a real-world 3D HOI annotation pipeline that provides a paradigm for obtaining rich 3D human-object interaction data from unlimited 2D images. Our proposed annotation process relies on the ability of 3D human and object reconstruction tools. In the future, with more advanced 3D-AIGC tools, the annotation efficiency will be largely improved. Moreover, our LLM-based 3d testing tasks proved that existing 3D general models are poor at 3D HOI understanding. As understanding 3D HOI is an important task, it requires more fine-annotated data to drive more general and capable models in the future.

\section{Conclusion}

In this work, we propose a method for annotating 3D HOIs from open-world single-view images and create Open3DHOI. The rich annotations in our dataset can support various 3D action tasks. 
For 3D HOI reconstruction, we introduce a 3D Gaussian optimizer that surpasses baselines. 
Results of current methods reveal that they are not yet capable of understanding 3D HOIs well.
We believe Open3DHOI will pave the way for future 3D HOI learning.

\section{Acknowledgment}
This work is supported in part by the National Natural Science Foundation of China under Grant 
No.62306175, 
62302296, 
72192821,
62472282.

\appendix

\section*{Appendix Overview}
The contents of this supplementary material are:
Sec.~\ref{sec:characteristics}: Characteristics of Open3DHOI.

Sec.~\ref{sec:method-detail}: Method Details.

Sec.~\ref{sec:results}: Additional Experiments.

\section{Characteristics of Open3DHOI}
\label{sec:characteristics}

\subsection{Image Selection for Open3DHOI}
Considering the complexity and difficulty of the 3D HOI annotation process, we only select images with single-person annotation from the existing 2D HOI dataset, HAKE, and SWIG-HOI. There are 63 images in our final dataset that have multiple objects interacting with one person. For these images, we split the annotation to keep one image having one HOI pair. 

\textbf{Interaction}. 
Notice that we have 3,671 interactions, more than our image number, 2,561, because one person can interact with an object with multiple actions, like drinking with and holding a bottle at the same time. Fig.~\ref{fig:action} shows the co-occurrence between the major object categories and actions, and Tab.~\ref{tab:obj_list} shows the object list in our Open3DHOI dataset. 

\textbf{Object size}.
The object size in our dataset varies significantly across different categories, and even within the same category, there is also a variation in size. In Fig.~\ref{fig:size_violin}, we chose object categories with more than 30 images and draw the size distribution in each category. We use the volume function from Trimesh to compute the volume of each object mesh, then take the cube root to obtain the size. We can see that object like elephants has larger sizes and bottles has smaller sizes. What's more, for objects like wine glasses, the size variation within the category is minimal, while for objects like couches, the variation is much larger. Fig.~\ref{fig:size_distribution} shows the size distribution of all images.

\textbf{Abnormal HOI}.
Because our dataset is created from 2D HOI datasets, which have many abnormal HOIs like standing on a chair, our dataset also contains many abnormal interactions. Fig.~\ref{fig:abnormal} shows some cases of our abnormal HOIs.

\begin{figure}[h]
    \centering
    \includegraphics[width=\linewidth]{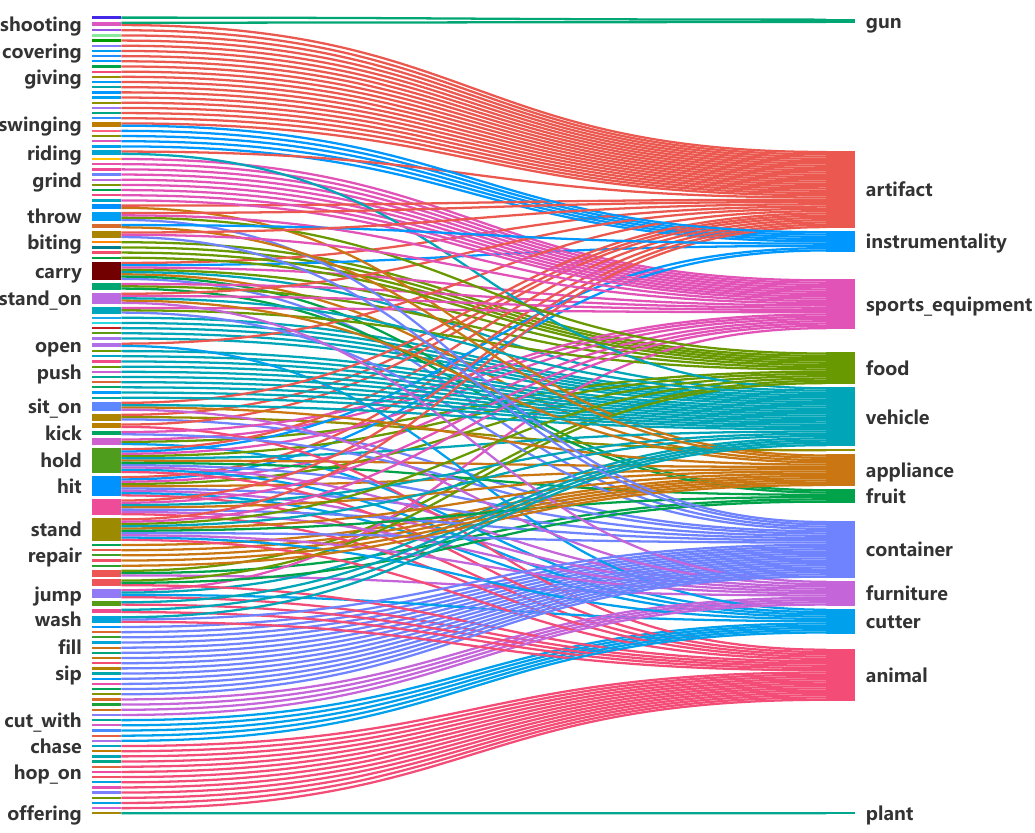}
    \caption{Co-occurence between major object category and actions in Open3DHOI.}
    \label{fig:action}
\end{figure}
\begin{figure}[h]
    \centering
    \includegraphics[width=\linewidth]{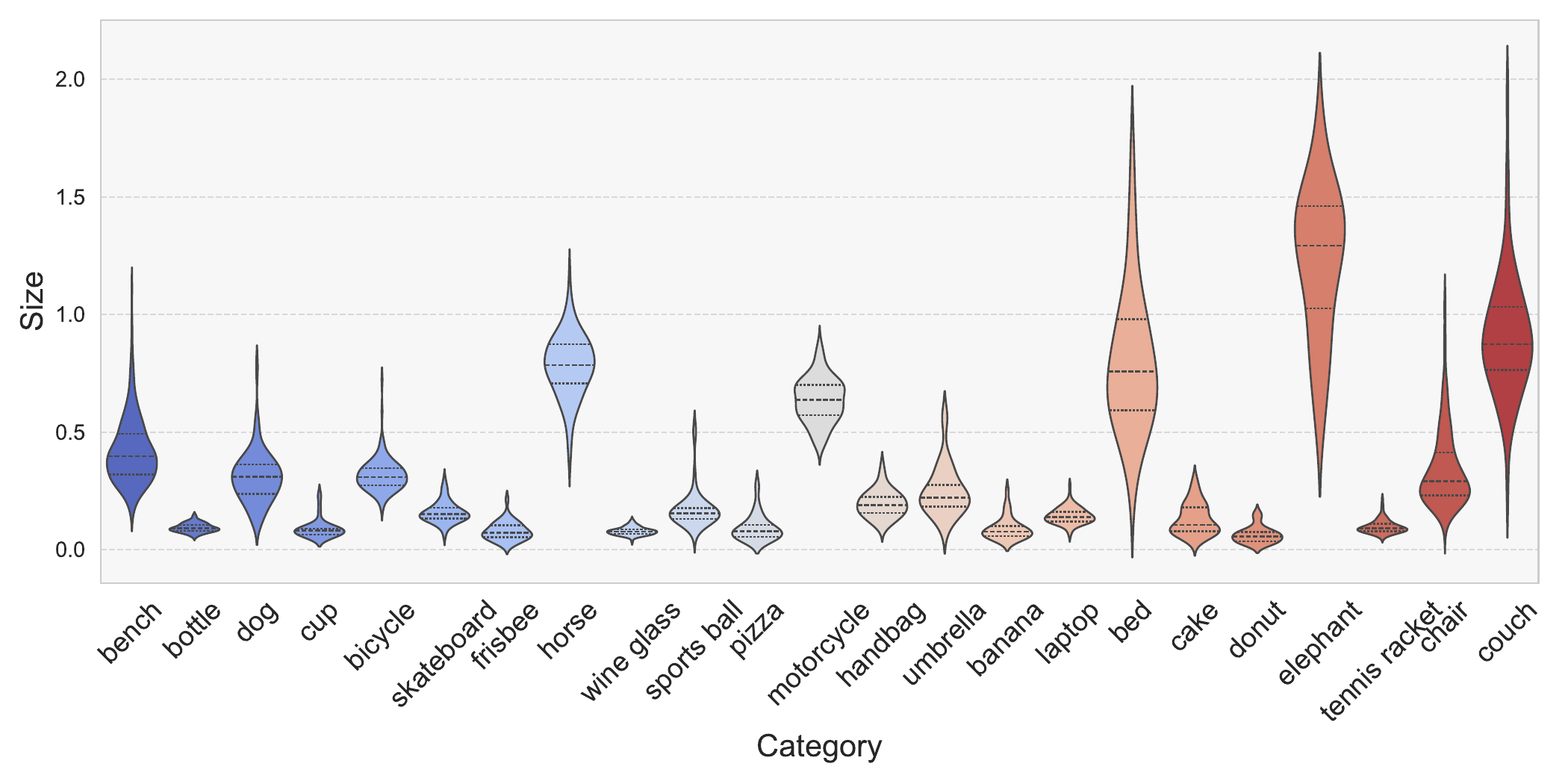}
    \caption{Object size distribution in different object categories.}
    \label{fig:size_violin}
\end{figure}
\begin{figure}[h]
    \centering
    \includegraphics[width=\linewidth]{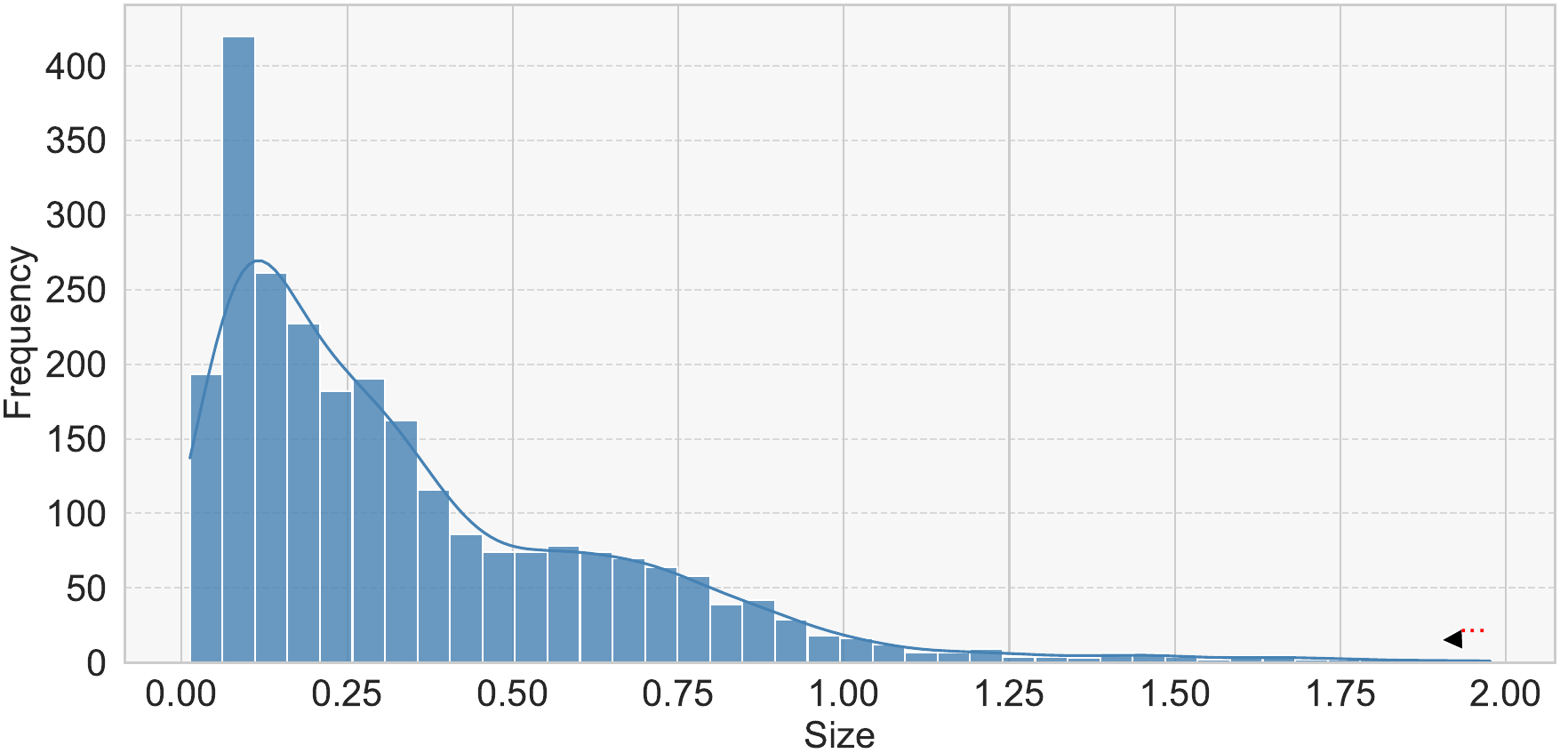}
    \caption{Object size distribution of all images.}
    \label{fig:size_distribution}
\end{figure}
\begin{figure}[h]
    \centering
    \includegraphics[width=\linewidth]{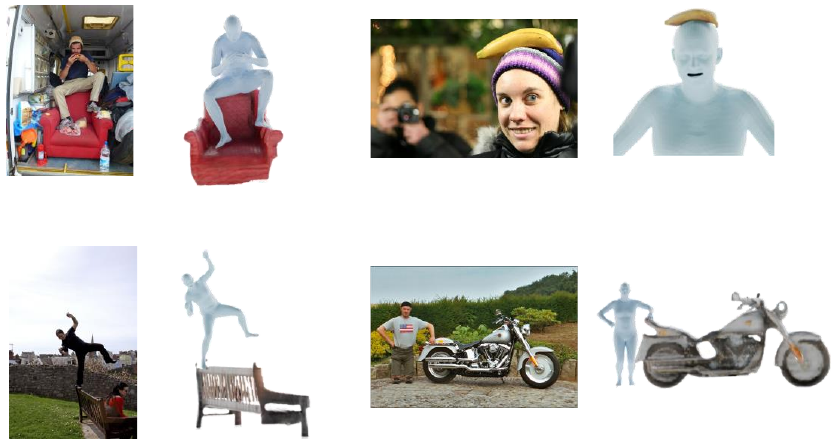}
    \caption{Abnormal HOIs in Open3DHOI.}
    \label{fig:abnormal}
\end{figure}

\begin{table*}[h]
\caption{Object categories in our Open3DHOI dataset.}
\label{tab:obj_list}
\resizebox{0.9\linewidth}{!}{
\begin{tabular}{c|c|c|c|c|c|c|c}
\toprule
Object Id & Object Class& Object Id & Object Class& Object Id & Object Class& Object Id & Object Class \\ \hline
0 & bird & 1 & television & 2 & surfboard & 3 & dining table \\ \hline
4 & mug & 5 & bench & 6 & goat & 7 & Gallus gallus \\ \hline
8 & fish & 9 & eggs & 10 & torch & 11 & rose \\ \hline
12 & award & 13 & guitar & 14 & pistol & 15 & ashcan \\ \hline
16 & baseball glove & 17 & bowl & 18 & shovel & 19 & bottle \\ \hline
20 & cookie & 21 & piano & 22 & home plate & 23 & furniture \\ \hline
24 & barrow & 25 & dog & 26 & boot & 27 & pot \\ \hline
28 & handcart & 29 & cell phone & 30 & donkey & 31 & hair drier \\ \hline
32 & basket & 33 & airplane & 34 & chain & 35 & oven \\ \hline
36 & box & 37 & cup & 38 & truck & 39 & bicycle \\ \hline
40 & snowboard & 41 & bucket & 42 & cat & 43 & pump \\ \hline
44 & hammock & 45 & skateboard & 46 & stone & 47 & sniper rifle \\ \hline
48 & cattle & 49 & tiger & 50 & power drill & 51 & mouse \\ \hline
52 & frisbee & 53 & helmet & 54 & violin & 55 & hobby \\ \hline
56 & car & 57 & book & 58 & horse & 59 & camel \\ \hline
60 & fire hydrant & 61 & backpack & 62 & backhoe & 63 & wine glass \\ \hline
64 & sports ball & 65 & clock & 66 & scissors & 67 & pizza \\ \hline
68 & raft & 69 & motorcycle & 70 & hammer & 71 & loaf of bread \\ \hline
72 & handbag & 73 & teddy bear & 74 & suitcase & 75 & vacuum cleaner \\ \hline
76 & pitcher & 77 & tie & 78 & vase & 79 & keyboard \\ \hline
80 & pumpkin & 81 & ice cream & 82 & boat & 83 & kite \\ \hline
84 & tarpaulin & 85 & umbrella & 86 & dinghy & 87 & package \\ \hline
88 & coffee cup & 89 & banana & 90 & laptop & 91 & knife \\ \hline
92 & mortar & 93 & hot dog & 94 & hairbrush & 95 & bed \\ \hline
96 & float & 97 & spoon & 98 & cow & 99 & cake \\ \hline
100 & sandwich & 101 & pen & 102 & bouquet & 103 & hoe \\ \hline
104 & jeep & 105 & lion & 106 & donut & 107 & apple \\ \hline
108 & whip & 109 & toilet & 110 & elephant & 111 & wrench \\ \hline
112 & tennis racket & 113 & liquor & 114 & hand glass & 115 & tricycle \\ \hline
116 & remote & 117 & bullet & 118 & pipage & 119 & baggage \\ \hline
120 & toothbrush & 121 & skis & 122 & chair & 123 & couch \\ \hline
124 & sculpture & 125 & fork & 126 & air cushion & 127 & light bulb \\ \hline
128 & sheep & 129 & pottery & 130 & carrot & 131 & barrel \\ \hline
132 & fire extinguisher &  &  &  &  &  &  \\ \bottomrule
\end{tabular}}
\end{table*}

\subsection{Contact Annotation}
\begin{figure}[h]
    \centering
    \includegraphics[width=\linewidth]{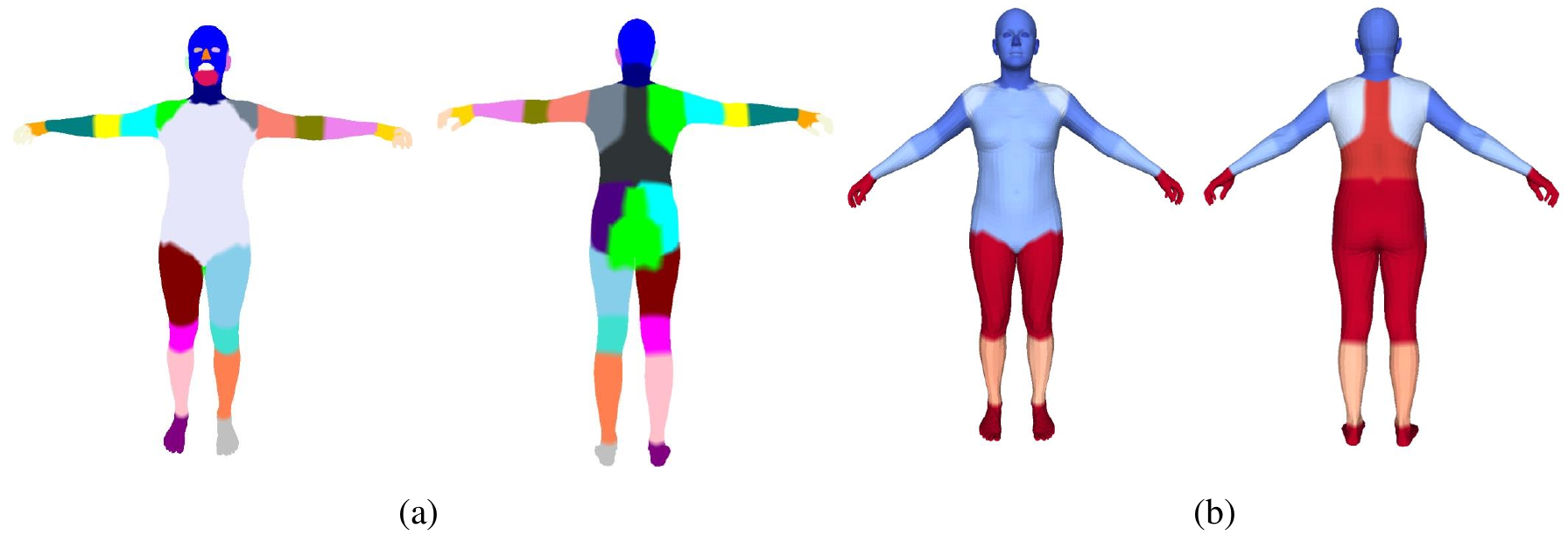}
    \caption{Body parts and annotation heat map.}
    \label{fig:contact_region}
\end{figure}

In our manual annotation process, we annotate the contact regions for images with qualified reconstruction. We split the human SMPL-X body into 34 parts and counted the number of annotations for each body part in Tab.~\ref{tab:body part}. In Fig.~\ref{fig:contact_region}, we show body parts on SMPL-X mesh and annotation heat map. It can be observed that interactions involving the hands, feet, and legs occur more frequently than those involving other body regions.

\begin{table}[t]
\caption{Body part name and annotation number.}
\label{tab:body part}
\begin{tabular}{l|l||l|l}
\hline
Body Part & Number &  Body Part& Number \\ \hline
bottom & 974 & head & 31\\
left elbow &55&left foot&335 \\
left palm& 689 & left hip & 435\\
left knee & 383 & left lower arm & 87 \\
left lower leg&27 & left shoulder & 112\\
left upper arm & 28 & left upper leg & 801\\
neck & 15 & right elbow & 58\\
right foot&332&right palm & 849\\
right hip & 417& right knee&361\\
right lower arm & 81 & right lower leg&195\\
right shoulder & 118& right upper arm& 37\\
right upper leg& 781& torso&76 \\
left eye& 1& right eye& 1\\
left fingers& 866 & right fingers&1065\\
left ear&1 &right ear&0\\
jaw& 22& nose &0\\
mouse& 32& back & 270\\
\hline
\end{tabular}
\end{table}

\section{Method Details}
\label{sec:method-detail}

\subsection{Coarse Reconstruction}
\label{sec:coarse recon}
In paper Fig.2, we introduce the process of coarse reconstruction. In this section, we provide additional details about this process. After reconstructing human and object meshes, we use depth to initialize coarse spatial alignment. We use Zoedepth to estimate depth information for each image and convert the depth to a point cloud $S$. We use an object mask to segment points of objects and place the object mesh to the point cloud center as ${Obj}_{init}$. Next, we use \cref{alg:alignment} to align the object mesh with the human mesh. 
\begin{algorithm}[h!]
\caption{Align object mesh with human mesh.}
\label{alg:alignment}
\textbf{Input: } Points cloud of scene $S$, 3D human points $H\_3D$, camera intrinsic parameter $K_h$, 3D object model ${Obj}_{init}$\\
\textbf{Output: } 3D points of objects $Obj\_3D$
\begin{algorithmic}
   \State 1. $H\_proj \gets $ project $H\_3D$ by $K_h=\left(\begin{array}{ccc}
        f & 0 & c_x \\
        0 & f & c_y \\
        0 & 0 & 1
        \end{array}\right)$
   \State $x_h^{proj} \gets \frac{\overline{z}c_x}{f}, y_h^{proj} \gets \frac{\overline{z}c_y}{f}$
   \State 
   \State 2. ${Index}_h \gets $ compare image and $H\_proj$ to obtain the indices of $S$ corresponding to $H\_3D$
   \State $H\_3D \gets H\_3D[ argsort(H\_3D[:,2])//2] $ get half of human points by depth
   \State $S_h \gets $ extract points belong to human in $S$ by ${Index}_h$
   \State 
   \State 3. $Scale, Translation \gets $ compare $S_h$ and $H\_3D$ to get 3D transformation
   \State $scale \gets $ using $\frac{1}{{N}^2} \sum_{i=1}^{N} \sum_{j=1}^{N} \| {p}_i - {p}_j \|_2$ to get scale of $S_h$ and $H\_3D$, and scale is $s_{H\_3D}/s_{S_h}$
   \State $translation \gets mean(S_h)-mean(H\_3D*scale)$
   \State
   \State 4. $Obj\_3D \gets $ operate ${Obj}_{init}$ by $Scale*{Obj}_{init}+ Translation$
   \State \textbf{return} $Obj\_3D$\;
\end{algorithmic}
\end{algorithm}

\subsection{Annotation Tools}
\subsubsection{Filtering Tool}
\begin{figure}[h]
    \centering
    \includegraphics[width=\linewidth]{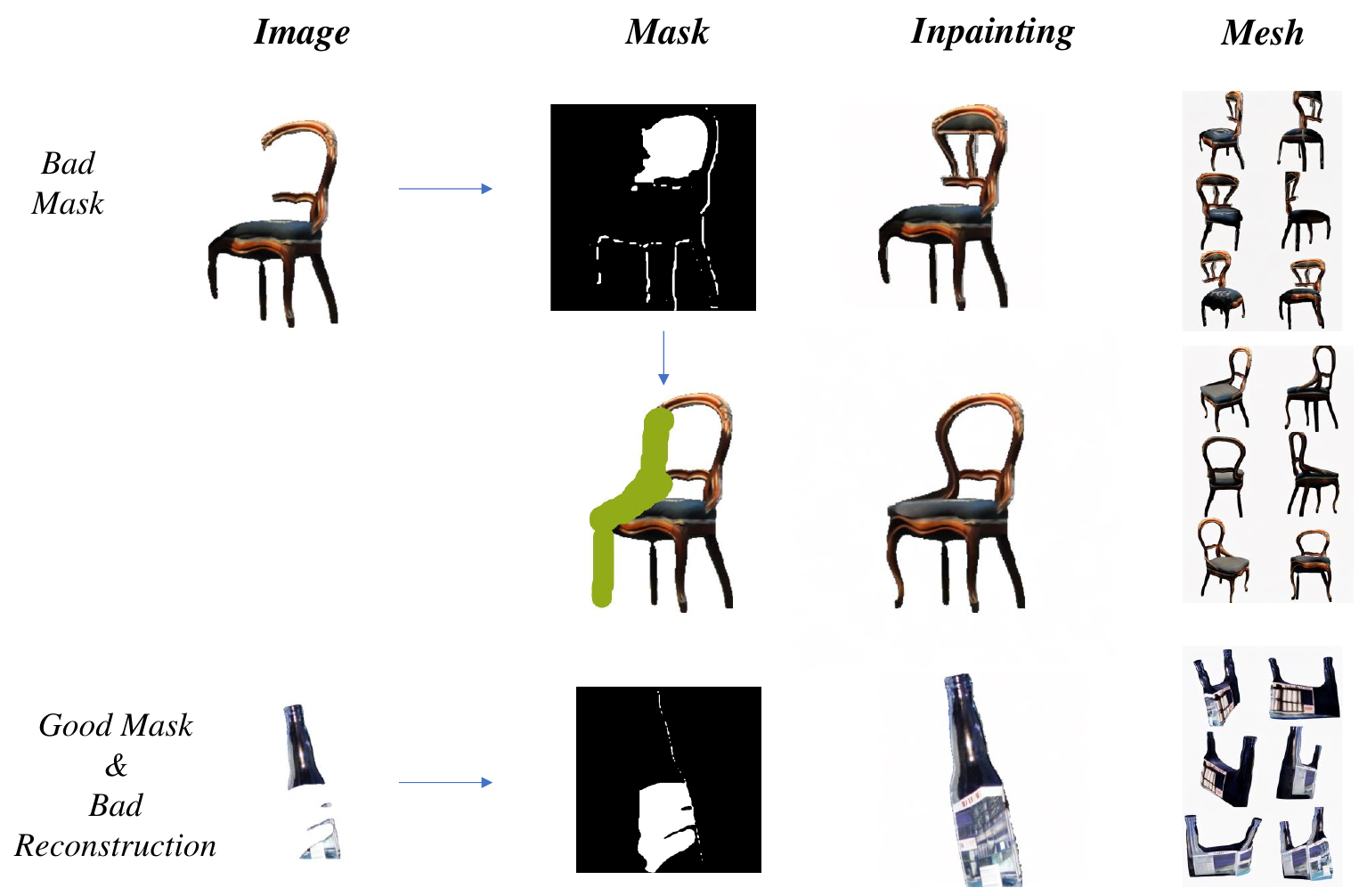}
    \caption{Special cases in filtering process.}
    \label{fig:filter_case}
\end{figure}

\begin{figure*}[h]
    \centering
    \includegraphics[width=\linewidth]{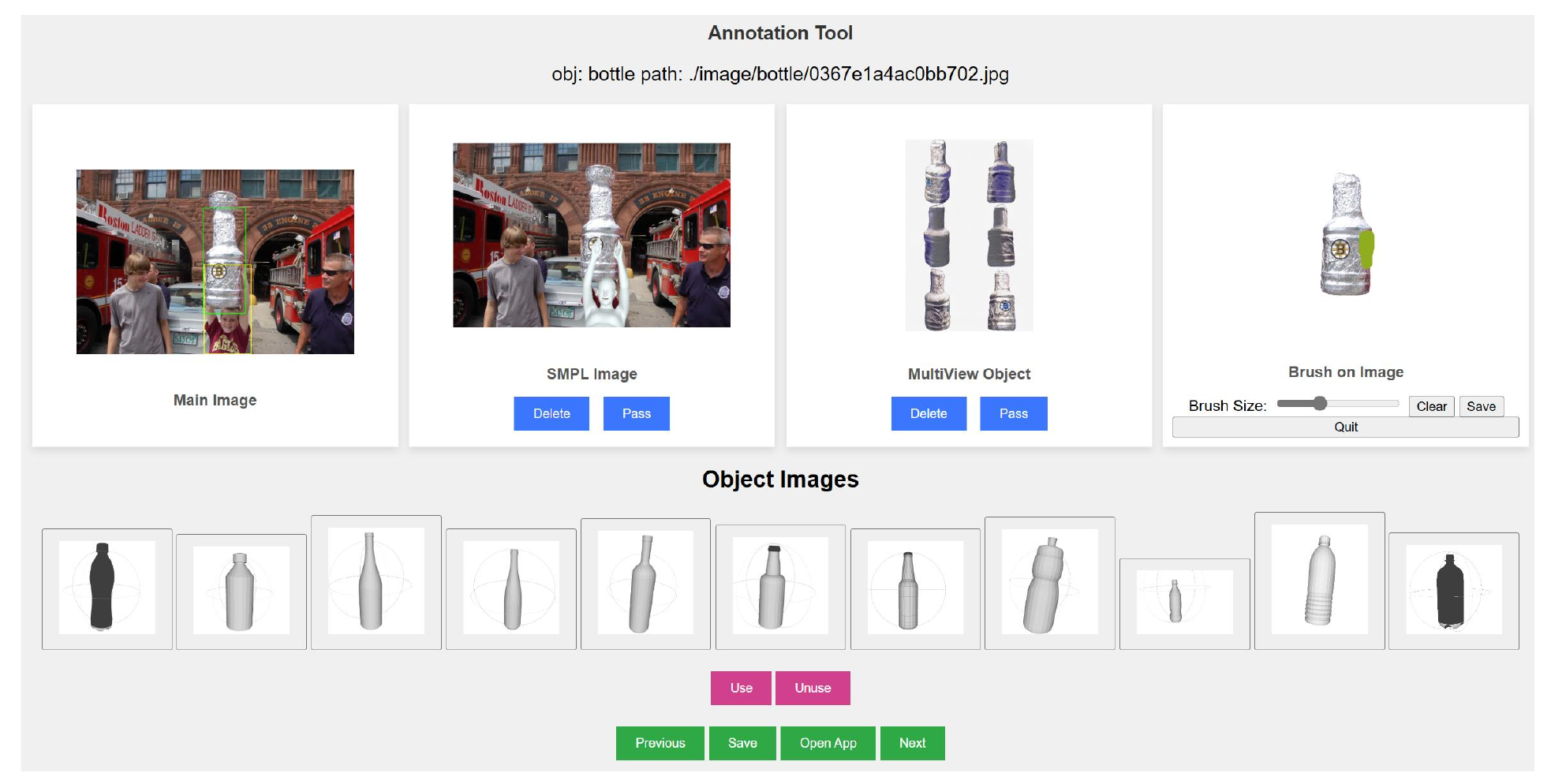}
    \caption{ Filtering tool.}
    \label{fig:filtering}
\end{figure*}

\begin{figure}[h]
    \centering
    \includegraphics[width=\linewidth]{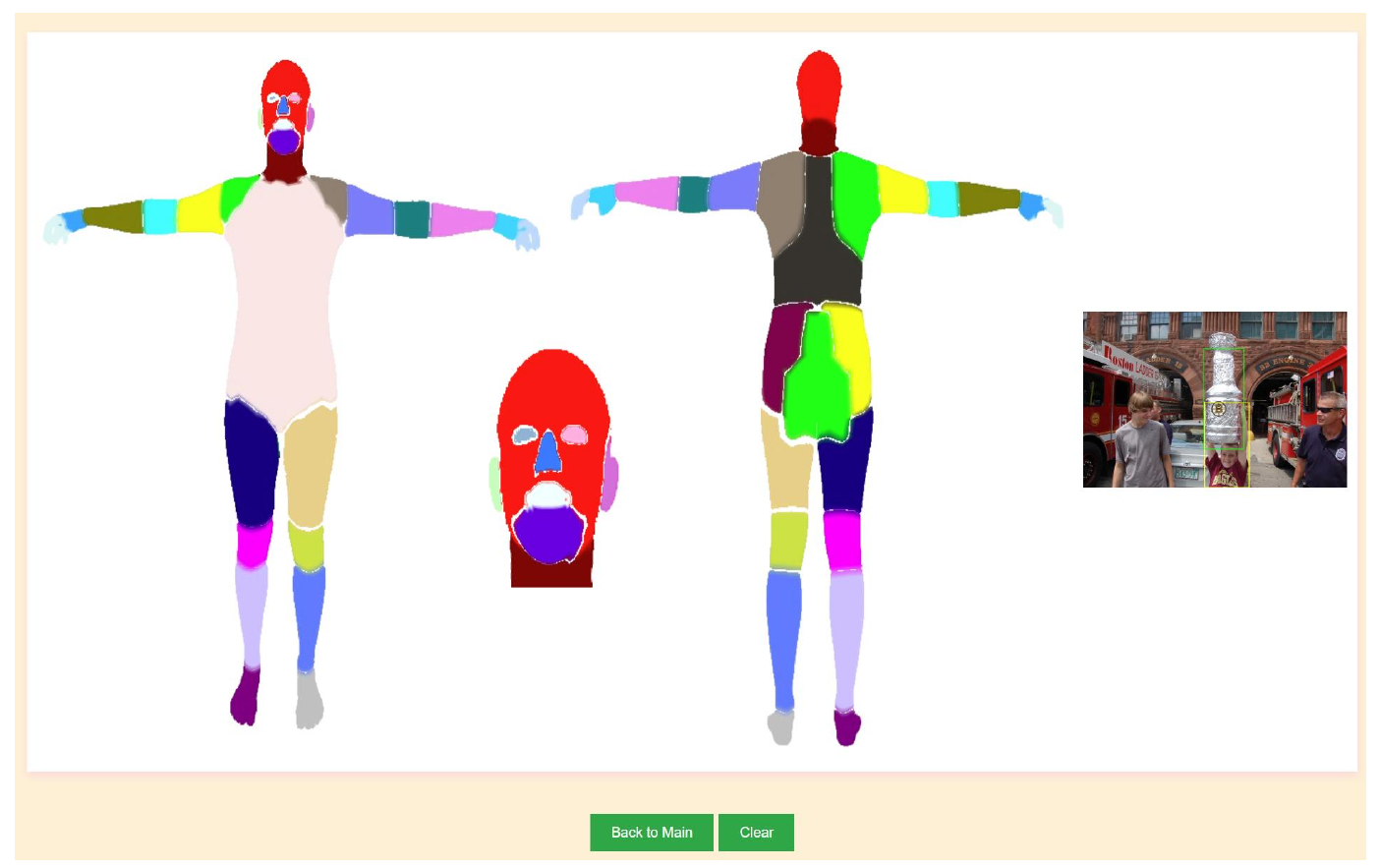}
    \caption{Contact part annotation tool.}
    \label{fig:contact_annot}
\end{figure}

Fig.~\ref{fig:filtering} (a) shows our filtering tool. 
First, we judge whether human reconstruction is qualified using the rendered image. There are two buttons, ``Delete'' and ``Pass'', if human reconstruction is bad, we click on the ``Delete'' button to delete this image otherwise we click on the ``Pass'' button and go to the next procedure to judge object reconstruction quality. According to the six-view rendering, we choose to keep the image and not. If the reconstruction is bad because the mask completion doesn't work well, we will ask the volunteer to correct the mask in the last column using a mouse brush. If the mask completion is not bad but the reconstruction is still terrible, or if the occlusion is too serious to reconstruct, we choose to click on the ``Delete'' button to delete this image. If the volunteer clicks on the ``Pass'' button for both human and object, then he needs to click on the ``Open App'' button on the bottom to open the contact annotation app in Fig.~\ref{fig:contact_annot}. Each body part in the app is clickable for volunteers to choose the contact part. After selection, the volunteer needs to go back to the main page and save the final annotation result. Fig.~\ref{fig:filter_case} shows cases with bad masks and with good masks but bad reconstructions. 


\subsubsection{3D Interaction Tool}
\textbf{Blender Annotation Tool}. When we have filtered human and object meshes, then we use the coarse reconstruction method in Sec.~\ref{sec:coarse recon} to initialize 3D HOI.
We designed a blender add-on for 3D HOI annotation. There are three buttons on the top, ``Load Meshes'', ``Export Object Pose and Location'' and ``Save Delete and Load Next''. The first is to load human and object meshes and image references. Volunteers need to adjust the objects’ positions, rotations, and scales using a mouse, while the human is fixed. After annotating, volunteers can use the second button to save the result and load the next image, or choose to use the third button to delete this image if it is hard to annotate.

\textbf{Fine Annotation Tool}. During the annotation process in Blender, the images were used as references without precise alignment. Although we ensured reasonable 3D interaction during the Blender annotation process, some objects' poses still exhibit discrepancies compared to the images. Fig.~\ref{fig:fine_3d} shows our 3D fine annotation tool based on ImageNet3D~\cite{imagenet3d}, to optimize the results from previous annotation. We select 581 images with IoU between human-object projection and mask lower than 0.5 and project a line set of meshes on the image. To ensure 3D interaction accuracy, we also project the meshes from three novel views. Volunteers need to click on the buttons to move, rescale, and rotate the object until it is aligned with the image.
\begin{figure*}[h]
    \centering
    \includegraphics[width=0.8\linewidth]{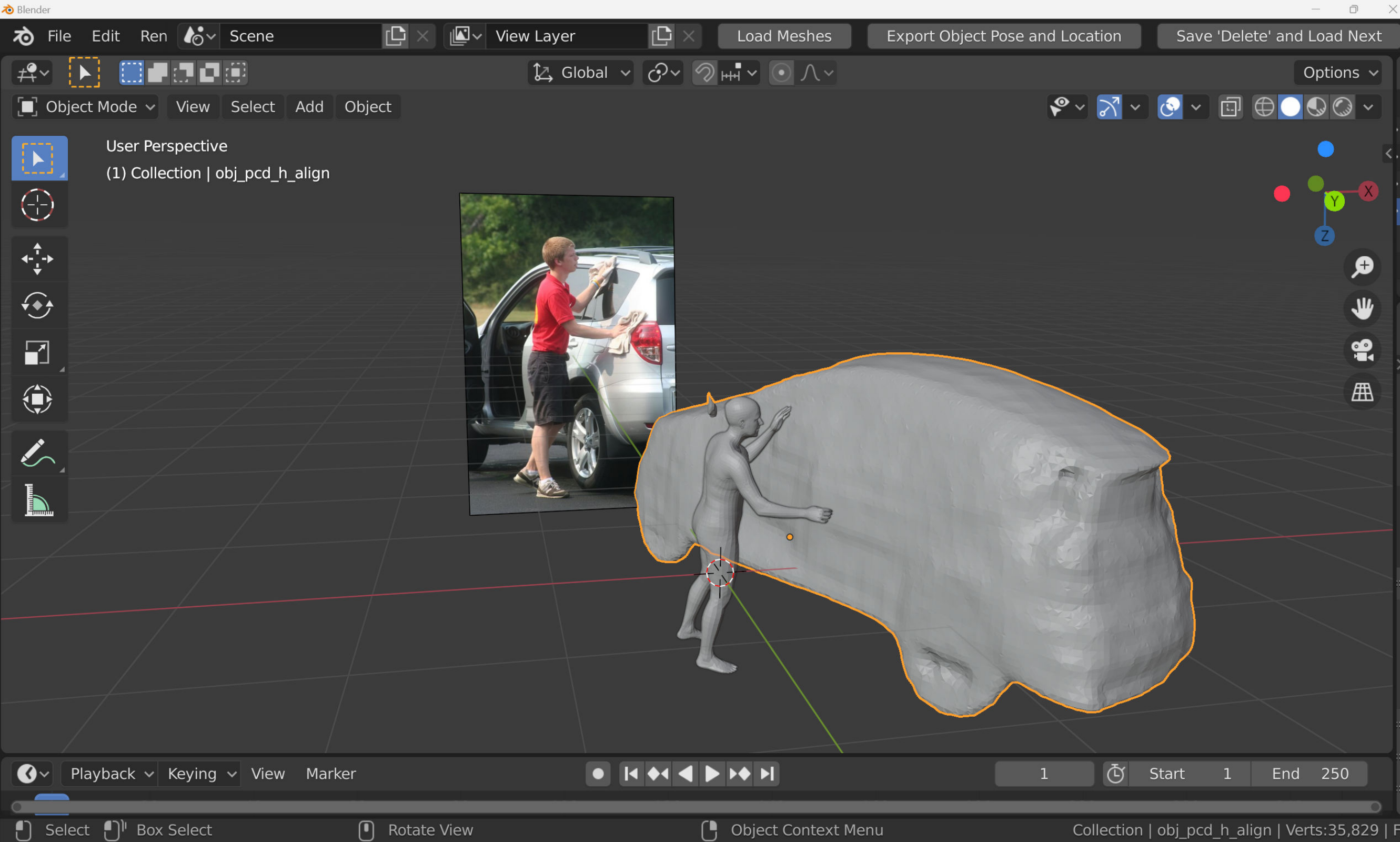}
    \caption{}
    \label{fig:action}
\end{figure*}

\begin{figure*}[h]
    \centering
    \includegraphics[width=0.95\linewidth]{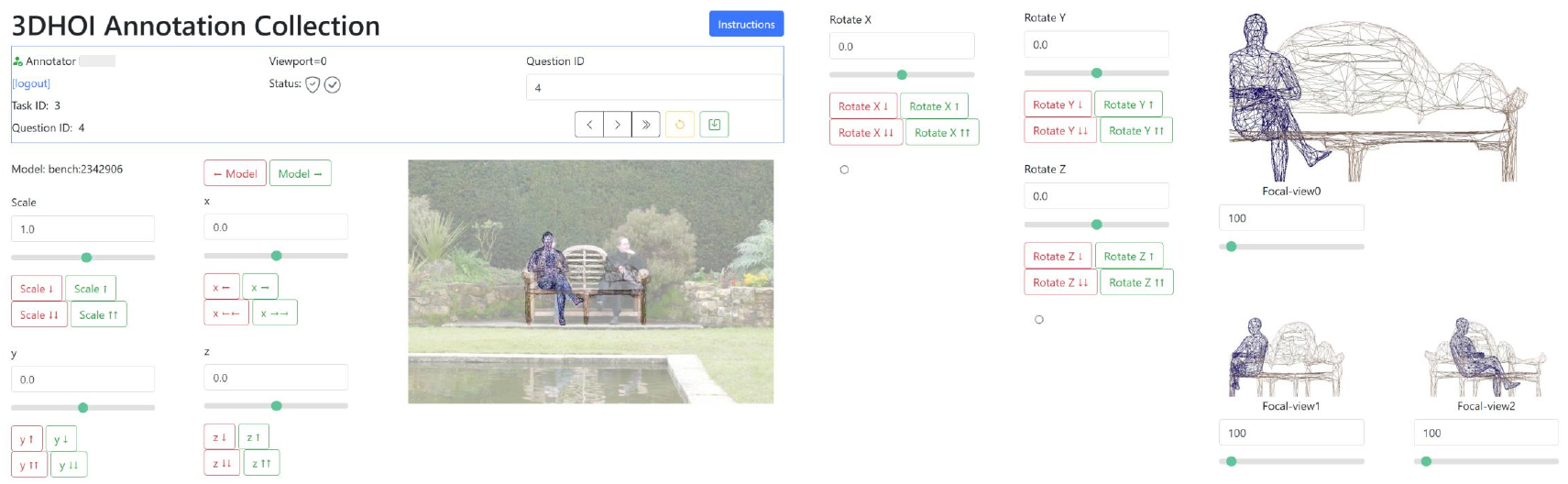}
    \caption{3D fine annotation tool.}
    \label{fig:fine_3d}
\end{figure*}
\subsubsection{Dataset quality}
\textbf{1)} \textbf{Human-object penetration rate}: we tested the penetration metric following\cite{multiperson}
by adding human-in-object penetration and object-in-human penetration together, which is 3.26 while PHOSA is 4.26. Notice that only considering penetration is not fair because in some cases where objects and humans are far away from each other also have zero penetration. Since we annotated human contact parts, so we also tested the distance between the annotated contact part and the  object divided by object size, the score of GT after normalization is 0.058, and PHOSA is 0.326. 
\textbf{2)} \textbf{Human and object projection error}: the human projection IoU is 0.621, the object projection error is 0.384, and the H+O projection error is 0.634. Notice that there is a significant occlusion of objects and humans in wild images, especially for objects, so this score can only serve as a reference. 
\textbf{3)} \textbf{Reconstruction quality}: since there is no GT object in our dataset, it is difficult to evaluate the quality of object reconstruction using traditional metrics. We use the inpainted GT object images and the projections of the annotated object mesh to compute SSIM and LPIPS for evaluation. Due to discrepancies between the object pose and the GT image, as well as the inherent differences between real images and mesh projections, including lighting, noise, \textit{etc.}, this evaluation is not entirely fair. However, our reconstruction scores still reached an LPIPS of 0.714 and an SSIM of 0.294, demonstrating that the quality of our reconstruction is high.
\begin{figure}[h]
    \centering
    \includegraphics[width=0.9\linewidth]{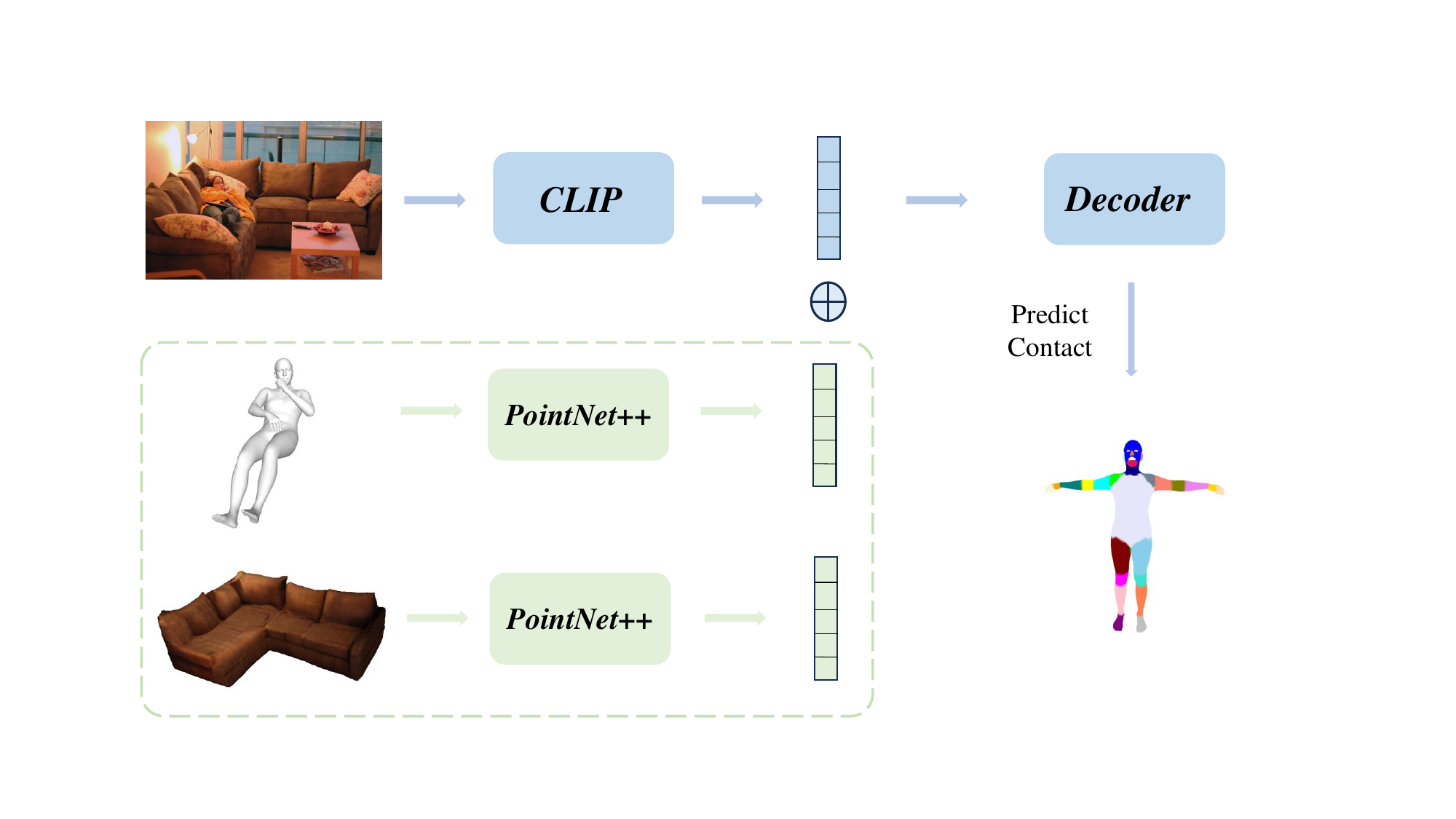}
    \caption{Our contact evaluation model.}
    \label{fig:contact_evaluation}
\end{figure}

\begin{figure*}[t]
    \centering
    \includegraphics[width=0.85\linewidth]{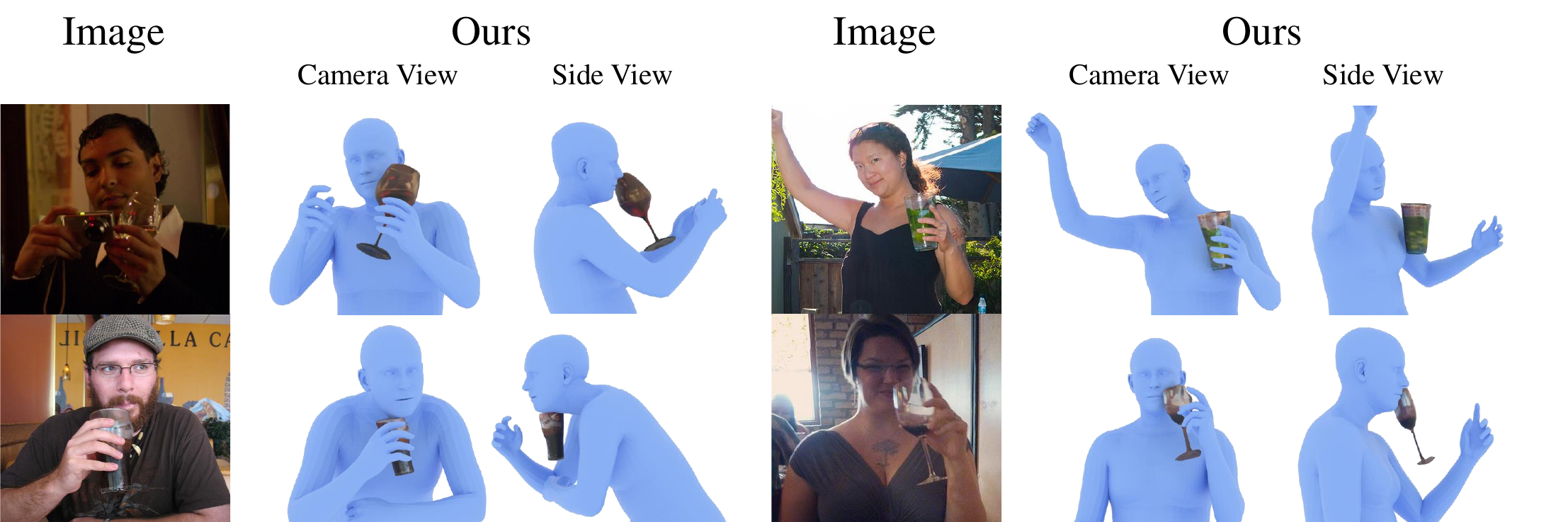}
    \caption{Failure cases of HOI-Gaussian.}
    \label{fig:failure}
\end{figure*}
\label{sec:results}
\subsubsection{Discussion}
Throughout the whole annotation process, we collected 2.5k+ images from 15k source images, resulting in a pass rate of 17\%, which indicates that most 2D HOI images are hard to reconstruct 3D representations. In the future, the filtering process can be accelerated by training a model to judge the reconstruction result, and volunteers only need to filter based on the predictions. Our annotation process has provided enough data to train a judge model. 

At the bottom of our filtering app, there are many object template buttons, which are designed to assign corresponding templates for images that closely match the template but have poor reconstruction quality. We build a template library for \textbf{58} object categories and totally \textbf{212} templates. Although we didn't use this library to build our Open3DHOI dataset, it is still very useful for future work. 

After our 3D fine annotation process, the IoU between human-object projection and mask increased from 0.48 to 0.57, and the IoU between object mesh projection and object mask increased from 0.32 to 0.48, which indicates that our fine annotation tool indeed improved the pose alignment.

\begin{table}[t]
\caption{Results of our contact evaluation task.}
\label{tab:contact_eval}
\resizebox{\linewidth}{!}{
\begin{tabular}{lccc}
\hline
Methods & Micro F1-Score \textuparrow& Hamming Loss \textdownarrow & Jaccard Index \textuparrow\\ \hline
2D & 0.6118 & 0.0874 &  0.4303 \\ 
2D\&3D & \textbf{0.6207}& \textbf{0.0844} & \textbf{0.4561} \\\hline
\end{tabular}}

\end{table}

\vspace{-5px}
\subsection{LLM Task Setting}
\subsubsection{PointLLM}
We used PointLLM-7B as a test model, and input our annotated human and object mesh vertices. Object vertices have colors and human vertices are colored black. When asking, we will tell PointLLM that ``The point cloud is a person interacting with an object. The person is black.'' first and then asks specific questions. To decrease the difficulty, we ask PointLM to generate a description first and use Qwen2.5~\cite{qwen2.5} to extract the exact word from our action and object list.

\subsubsection{ChatPose}
In Sec.6.2, we state that we select images with multiple images. Although our dataset only contains single-person annotation, there are still many images with more than one person, we used Detectron2~\cite{detectron2} to detect these images for our testing. Our task is to ask ChatPose to locate the specific person interacting with the specific object according to its understanding of the interaction in the image. The pose it answered has no root pose and location, so we compare the prediction with GT using the same root pose, zero pose at zero location. The metrics we used are MPJPE (Mean Per Joint Position Error) and MPVPE (Mean Per Vertex Position Error), which are common metrics in human pose estimation.
\section{Additional Experiments}

\subsection{Contact Evaluation}
Since our dataset contains contact annotations, we want to evaluate whether 3D information would be conducive to estimating contact regions compared to image only. Therefore, we design a simple pipeline to estimate the contact regions. As Fig.~\ref{fig:contact_evaluation} shows, we use clip-ViT-B/32 to encode image and pointnet++ to encode normalized human point clouds and object point clouds respectively. Image features and point clouds features of human and object are fused and put into an MLP decoder. 
We treat this problem as a multi-label classification task and use Micro F1 Score, Hamming Loss and Jaccard Index to evaluate the accuracy. The Micro F1 Score calculates precision and recall globally across all labels. Hamming Loss measures the fraction of incorrect label predictions over the total number of labels. Jaccard Index evaluates the similarity between the predicted and true label sets for each sample.
Our current implementation simply concatenates 3D and 2D features and is trained on only 2,000 samples. However, our results over multiple metrics in Tab.~\ref{tab:contact_eval} still indicate that 3D information is beneficial to the estimation of contact regions.


\subsection{Failure Cases}

Fig.~\ref{fig:failure} shows some failure cases of our HOI-Gaussian optimizer. In these cases, human body parts occlude each other severely, and the object happens to be located between the occluded areas, which becomes challenging to determine which body part the object should contact with.

\subsection{More Results}

\begin{figure*}[h]
    \centering
    \includegraphics[width=\linewidth]{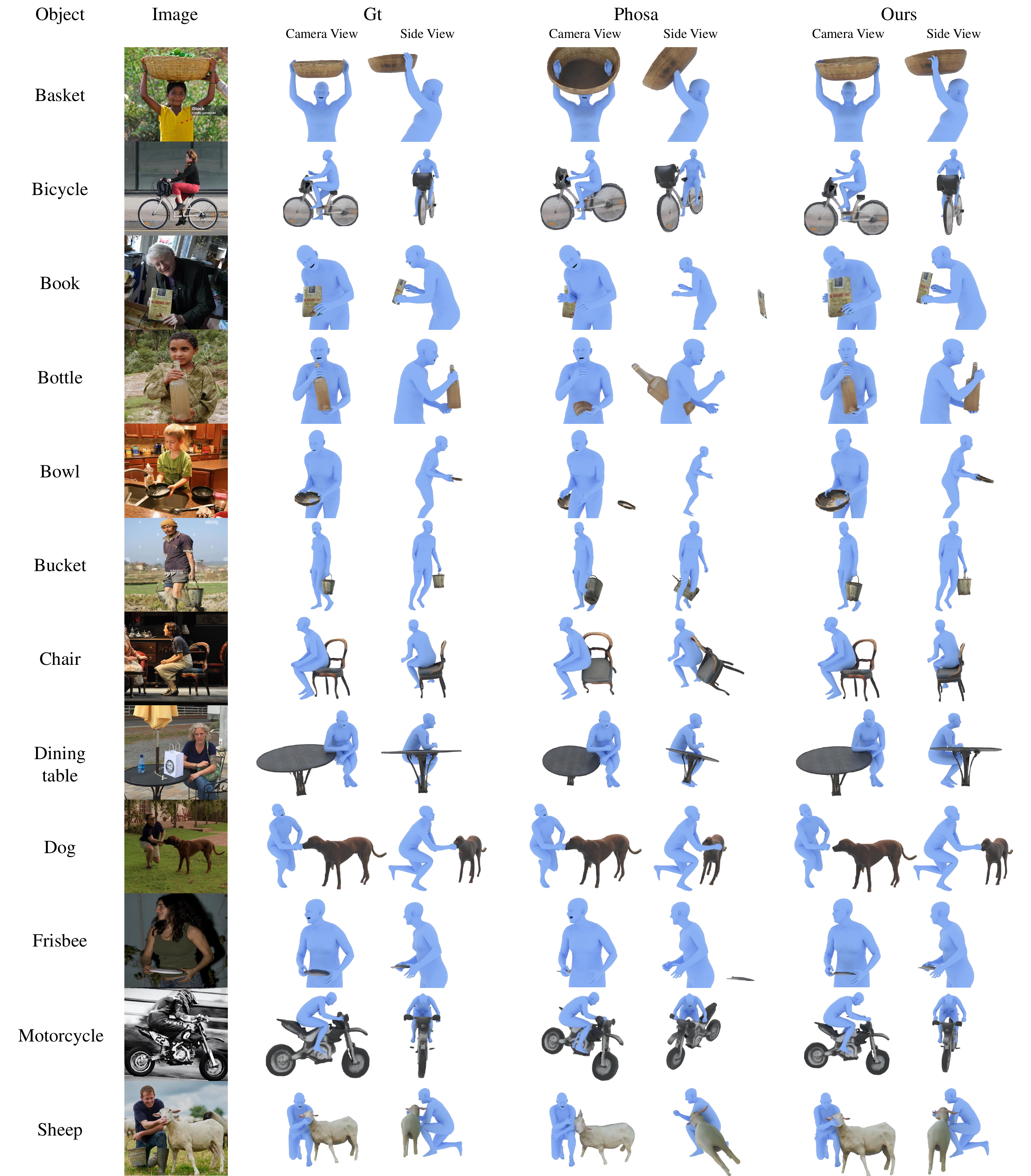}
    \caption{More results.}
    \label{fig:more_results}
\end{figure*}

Fig.~\ref{fig:more_results} shows more results comparison between GT, PHOSA, and Ours.

\bibliographystyle{ieeenat_fullname}
\bibliography{main}

\end{document}